%% file: main_jmlr.tex
\documentclass[twoside,11pt]{article}

\usepackage{subcaption}
\usepackage{booktabs}
\usepackage{url}
\usepackage{multirow}
\usepackage{tikz}
\usetikzlibrary{calc}
\usetikzlibrary{decorations.pathmorphing}
\usepackage{tikz}
\usetikzlibrary{calc}
\usetikzlibrary{decorations.pathmorphing}
\usetikzlibrary{patterns,arrows,automata,positioning,decorations,shapes,calc}

\tikzstyle{normalvertex}=[circle,fill=white,draw=black]
\tikzstyle{emptyvertex}=[draw,circle,minimum size=7pt,inner sep=0pt]
\tikzstyle{tinyvertex}=[draw,circle,minimum size=3pt,inner sep=0pt]
\tikzstyle{thickedge}=[draw,gray!60,line width=1.6pt,-]

\usepackage[abbrvbib]{jmlr2e}

\usepackage{pgfplots}
\usepgfplotslibrary{colorbrewer}

\pgfdeclarelayer{background}
\pgfsetlayers{background,main}

\usetikzlibrary{shapes}
\usetikzlibrary{arrows.meta,backgrounds,automata, chains}
\usetikzlibrary{positioning,shapes,matrix,calc}
\tikzstyle{vertex}=[circle, draw, fill=gray!80!white,thick,scale=1.2]
\tikzstyle{edge}=[draw=black, thick,-]

\usepackage{bm}

\definecolor{purple}{RGB}{147,7,204}
\definecolor{blue}{RGB}{10,153,201}
\definecolor{orange}{RGB}{254,128,41}
\definecolor{gray}{RGB}{239,240,241}
\definecolor{pink}{RGB}{254,15,127}
\definecolor{green}{RGB}{140,211,89}

\definecolor{color1}{RGB}{254,15,127}
\definecolor{color2}{RGB}{10,153,201}
\definecolor{color3}{RGB}{194,145,62}
\definecolor{color4}{RGB}{254,128,41}
\definecolor{color5}{RGB}{254,191,185}
\definecolor{color6}{RGB}{140,211,89}
\definecolor{color7}{RGB}{245,221,66}

\newcommand{\oms}{\{\!\! \{}
\newcommand{\cms}{\}\!\! \}}

\usepackage{csquotes}
\usepackage{paralist}

\usepackage{bm}

\definecolor{purple}{RGB}{147,7,204}
\definecolor{green}{RGB}{5,100,18}
\definecolor{blue}{RGB}{10,153,201}
\definecolor{orange}{RGB}{254,128,41}
\definecolor{gray}{RGB}{239,240,241}

\usepackage{thmtools}		
\usepackage{mleftright}
\usepackage{stmaryrd}
\usepackage{nicefrac}

\newtheorem{oproblem}{Open Challenge}

\usepackage{thm-restate}
\usepackage{mathtools}
\usepackage{fixmath}
\usepackage{siunitx}
\usepackage{color}

\usepackage{pifont}

\usepackage{setspace}
\usepackage{microtype}
\microtypecontext{spacing=nonfrench}
\usepackage{ellipsis}
\usepackage{xspace}

\usepackage{tcolorbox}

\makeatletter
\def\thmt@refnamewithcomma #1#2#3,#4,#5\@nil{%
	\@xa\def\csname\thmt@envname #1utorefname\endcsname{#3}%
	\ifcsname #2refname\endcsname
	\csname #2refname\expandafter\endcsname\expandafter{\thmt@envname}{#3}{#4}%
	\fi
}
\makeatother
\usepackage[capitalise,noabbrev]{cleveref}   

\newcommand{\new}[1]{\emph{#1}}

\newcommand{\cO}{\ensuremath{{\mathcal O}}\xspace}

\newcommand{\bbR}{\ensuremath{\mathbb{R}}}

\newcommand{\bbN}{\ensuremath{\mathbb{N}}}

\newcommand{\cnp}{\textsf{NP}\xspace}

\newcommand{\RR}{\mathbb{R}}

\newcommand{\NN}{\mathbb{N}}

\renewcommand{\vec}[1]{\mathbf{#1}}

\usepackage{lastpage}
\ShortHeadings{Weisfeiler and Leman go Machine Learning: The Story so far}{Morris, Lipman, Maron, Rieck, Kriege, Grohe, Fey, Borgwardt}
\firstpageno{1}

\jmlrheading{24}{2023}{1-\pageref{LastPage}}{3/22}{5/23}{22-0240}{Christopher Morris, Yaron Lipman, Haggai Maron, Bastian Rieck, Nils M. Kriege, Martin Grohe, Matthias Fey, Karsten Borgwardt}

\firstpageno{1}

\begin{document}

\title{Weisfeiler and Leman go Machine Learning: The Story so far}

\author{\name Christopher Morris  \email morris@cs.rwth-aachen.de \\
	\addr Department of Computer Science\\
	RWTH Aachen University\\
	Aachen, Germany
	\AND
	\name Yaron Lipman  \email yaron.lipman@weizmann.ac.il \\
	\addr Meta AI Research \\Department of Computer Science and Applied Mathematics\\
	Weizmann Institute of Science\\
	Rehovot, Israel
	\AND
	\name Haggai Maron \email hmaron@nvidia.com\\
	\addr NVIDIA Research\\
	Tel Aviv, Israel
	\AND
	\name Bastian Rieck  \email bastian.rieck@helmholtz-muenchen.de\\
	\addr AIDOS Lab, Institute of AI for Health\\
	Helmholtz Zentrum München and Technical University of Munich\\
	Munich, Germany
	\AND
	\name  Nils M.\ Kriege  \email nils.kriege@univie.ac.at \\
	\addr Faculty of Computer Science, University of Vienna, Vienna, Austria \\
	Research Network Data Science, University of Vienna, Vienna, Austria
	\AND
	\name Martin Grohe \email grohe@informatik.rwth-aachen.de\\
	\addr Department of Computer Science\\
	RWTH Aachen University\\
	Aachen, Germany
	\AND
	\name Matthias Fey \email matthias@kumo.ai\\
	\addr Kumo.AI\\
	Mountain View, CA
	\AND
	\name Karsten Borgwardt\thanks{Karsten Borgwardt is now at the Max Planck Institute of Biochemistry in Martinsried, Germany.} \email karsten.borgwardt@bsse.ethz.ch\\
	\addr Machine Learning \& Computational Biology Lab
	\\Department of Biosystems Science and Engineering\\
	ETH Zürich, Basel, Switzerland and\\ 
 Swiss Institute of Bioinformatics, Lausanne, Switzerland\\
}

\editor{David Wipf}

\maketitle

\newpage
\begin{abstract}% 
	In recent years, algorithms and neural architectures based on the Weisfeiler--Leman algorithm, a well-known heuristic for the graph isomorphism problem, have emerged as a powerful tool for machine learning with graphs and relational data. Here, we give a comprehensive overview of the algorithm's use in a machine-learning setting, focusing on the supervised regime. We discuss the theoretical background, show how to use it for supervised graph and node representation learning, discuss recent extensions, and outline the algorithm's connection to (permutation-)equivariant neural architectures. Moreover, we give an overview of current applications and future directions to stimulate further research.
\end{abstract}


\section{Introduction}\label{sec:intro}
\begin{figure}[t]
	\centering
	\includegraphics[width=\textwidth]{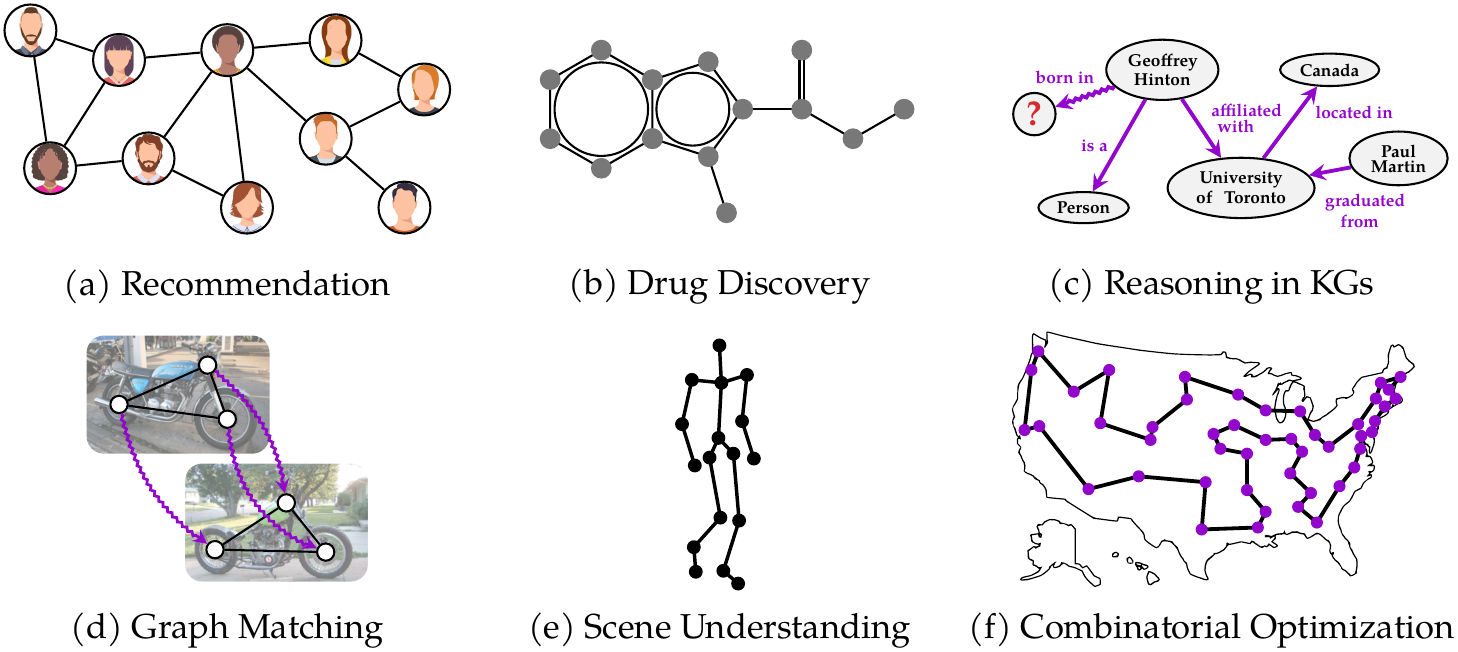}
	\caption{Example use cases for machine learning with graphs.}
	\label{fig:overview}
\end{figure}

Graph-structured data is ubiquitous across application domains, ranging from chemo- and bioinformatics~\citep{Barabasi2004,Jum+2021,Sto+2020} to computer vision~\citep{Sim+2017}, and social network analysis~\citep{Eas+2010}; see~\cref{fig:overview} for an overview of application areas. We need techniques exploiting the rich graph structure and feature information within nodes and edges to develop successful machine-learning models in these domains. Due to the highly non-regular structure of real-world graphs, most approaches first generate a vectorial representation of each graph or node, so-called \emph{node} or \new{graph embeddings}, respectively, to apply standard machine learning tools such as linear regression, random forests, or neural networks.

For successful (supervised) machine learning with graphs, node and graph embeddings need to address the following key challenges:

\begin{enumerate}
	\item The graph embedding needs to be \emph{invariant} to any permutation of the graph's nodes, i.e., the output of the graph embedding must not change for different orderings.
	\item In the case of node embeddings, the embedding needs to be \emph{\new{(permutation-)}equivariant} to node orderings, i.e., reordering of the input results in a reordering of the output, accordingly.
	\item The embeddings need to \new{scale} to large, real-world graphs and large sets thereof.
	\item The embeddings need to consider \emph{attribute} or \emph{label information}, e.g., real-valued vectors attached to nodes and edges.
	\item Finally, the embeddings need to \emph{generalize} to unseen instances and ideally easily adapt or be robust to changes in the data distributions.
\end{enumerate}

To address the above challenges, numerous approaches have been proposed in recent years---most notably, graph embedding approaches based on spectral techniques~\citep{Ath+2017,Lux+2008}, \new{graph kernels}~\citep{Borg+2020,Kriege2020b}, and neural approaches~\citep{Cha+2020,Gil+2017,Sca+2009} for both node and graph embeddings.
Here, graph kernels are positive semi-definite functions, expressing a pairwise similarity between graphs. Especially, graph kernels based on the \new{Weisfeiler--Leman algorithm}~\citep{Wei+1968}, a graph comparison algorithm originally developed to address the graph isomorphism problem, see below, and corresponding neural architectures, known as \new{graph neural networks} (GNNs), have recently advanced the state-of-the-art in (semi-)supervised node-level and graph-level machine learning.

The \new{($1$-dimensional)} Weisfeiler--Leman ($1$-WL)\footnote{
	We use the spelling ``Leman'' here as A.~Leman, co-inventor of the algorithm, preferred it over the transcription ``Lehman''; see
	\url{https://www.iti.zcu.cz/wl2018/pdf/leman.pdf}. If a paper used the spelling ``Lehman'' for a method's name, e.g., ``Weisfeiler--Lehman subtree kernel''~\citep{She+2009b}, we used it as well.

} or \new{color refinement} algorithm is a well-known heuristic for deciding whether two graphs are \new{isomorphic}, i.e., exactly match structure-wise. Given an initial \emph{coloring} or \emph{labeling} of the nodes of both graphs, e.g., their degree or application-specific information, in each iteration, two nodes with the same label get different labels if the number of neighbors carrying a particular label is not equal, see~\cref{fig:wlrelabel} for an illustration. If, after any iteration, the number of nodes annotated with a certain label is different in both graphs, the algorithm terminates, and we conclude that the two graphs are not isomorphic. This simple algorithm is already quite powerful in distinguishing non-isomorphic graphs~\citep{Bab+1979} and has been therefore applied in many areas, see, e.g.,~\citet{Gro+2014,Ker+2014,Zha+2017}, including graph classification~\citep{She+2011}. On the other hand, it is easy to see that the algorithm cannot distinguish all non-isomorphic graphs~\citep{Cai+1992}. For example, it cannot distinguish graphs with different triangle counts, see~\cref{fig:wlcounter}, or, in general, cyclic information~\citep{Arv+2015}, which is an important feature in social network analysis~\citep{Mil+2002,New2003} and chemical molecules. To increase the algorithm's expressive power, it has been generalized from labeling nodes to $k$-tuples, defined over the set of nodes, leading to a more powerful graph isomorphism heuristic, denoted \new{$k$-dimensional Weisfeiler--Leman} algorithm ($k$-WL). The $k$-WL was investigated in-depth by the theoretical computer science community, see, e.g.,~\citet{Cai+1992,Gro2017,Kie2020a}.

\citet{She+2009b} first used the $1$-WL as a graph kernel, the so-called \new{Weisfeiler--Lehman subtree kernel}. The kernel's idea is to compute the $1$-WL for a fixed number of steps, resulting in a color histogram or feature vector of color counts for each graph. Subsequently, taking the pairwise inner product between these vectors leads to a valid kernel function. Hence, the kernel measures the similarity between two graphs by counting common colors in all refinement steps. Similar approaches are popular in chemoinformatics for computing vectorial descriptors of chemical molecules~\citep{Rogers2010}.

Graph kernels were the primary approach for learning on graphs for several years, leading to new state-of-the-art results on many graph classification tasks. However, one limitation, in particular of the most efficient graph kernels, is that their feature vector representation corresponds to enumerating particular classes of subgraphs, and kernel computation corresponds to finding exactly matching pairs of these subgraphs in two graphs; thereby, partial similarities of subgraphs in two graphs may be missed. GNNs have emerged as a machine learning framework that aims to address these limitations. Primarily, they can be viewed as a neural version of the $1$-WL algorithm, where continuous feature vectors replace colors, and neural networks are used to aggregate over local node neighborhoods~\citep{Gil+2017,Ham+2017,Mor+2019}.

Recently, links between the two above paradigms emerged. \citet{Mor+2019,Xu+2018b} showed that any possible GNN architecture cannot be more powerful than the $1$-WL in terms of distinguishing non-isomorphic graphs. Moreover, this line of work has been extended by deriving more powerful neural architectures, based on the $k$-WL~\citep{Mar+2019b,Mar+2019,Mor+2019,Morris2020b}, subsequently shown to be universal~\citep{Azi+2020,keriven2019,Mar+2019b}, i.e., being able to represent any continuous, bounded, invariant or equivariant function over the set of graphs.

\subsection{Present Work}

In this paper, we survey the application of the Weisfeiler--Leman algorithm to machine learning with graphs. To this end, we overview the algorithm's theoretical properties and thoroughly survey graph kernel approaches based on the Weisfeiler--Leman paradigm. Subsequently, we also overview the recent progress in aligning the algorithm's expressive power with equivariant neural networks, showing $1$-WL's and $k$-WL's equivalence to GNNs and more powerful higher-order GNNs, respectively. Alongside, we also survey works proving universality results of invariant and equivariant neural architectures for graphs. Moreover, we review recent efforts extending GNNs' expressive power, their generalization abilities,w and exemplary applications of the algorithm's use in machine learning with graphs. Finally, we discuss open questions and future research directions.

\subsection{Related Work}

In the following, we briefly discuss related work relevant to the present survey.

\subsubsection{Graph Kernels}

Intuitively, a graph kernel is a function measuring the similarity of a pair of graphs, see~\cref{notation} for a formal definition and \citet{Moh+2012} as well as \citet{Sha+2014} for an introduction to kernel functions for machine learning. Graph kernels were the  dominant approach in machine learning for graphs, especially for graph classification with a relatively small number of graphs, for several years, see~\citet{Borg+2020} and \citet{Kriege2020b} for thorough surveys. Starting from the early 2000s, researchers proposed a plethora of graph kernels, e.g., based on  shortest-paths~\citep{Bor+2005a}, random walks~\citep{Gae+2003, Kan+2012, Kashima2003, Sug+2015, Kriege22}, small subgraphs~\citep{Shervashidze2009,Kri+2012}, local neighborhood information~\citep{Cos+2010,Mor+2017,Morris2020b,She+2011}, Laplacian information~\citep{Kon+2016}, and matchings~\citep{Fro+2005,Joh+2015,Kriege2016,Nik+2017,Woznica2010}.

\subsubsection{Molecule Descriptors in Cheminformatics}\label{sec:rw:mol}
The representation of small molecules by their structure to explain their chemical properties constitutes one of the early applications of graph theoretical concepts and has influenced modern graph theory~\citep{Biggs1986}. Cheminformatics applies computer science methods to analyze chemical data and comprises several graph-theoretical and machine-learning problems. Finding a unique representation of a molecular structure corresponds to the graph canonization problem. Using the experimentally obtained bioactivity data of small molecules to predict the activity of untested molecules to find promising drug candidates is an instance of a graph classification or regression task.
Therefore, it is not surprising that some techniques developed in cheminformatics are closely related to machine learning with graphs and the Weisfeiler--Leman method. We briefly review the developed methods and their relation to state-of-the-art techniques.

\paragraph{Morgan's Algorithm}
In 1965, \citet{Morgan1965} proposed a method to generate unique identifiers for molecules, up to isomorphism, see~\cref{notation}, that was implemented at the Chemical Abstracts Service\footnote{\url{www.cas.org}} to index and provide chemistry-related information. To this end, the atoms are numbered canonically based on the atom and bond types and their structure. To increase efficiency, ambiguities are reduced by computing, for each node $v$, its \emph{connectivity value} $\mathsf{ec}(v)$.
Let $G = (V,E)$ be a (molecular) graph, initially we assign $\mathsf{ec}^{(1)}(v)=\deg(v)$ to every node $v$ in $V$, where $\deg(v)$ is the degree of $v$. Then, the values $\mathsf{ec}^{(i)}(v)$ are computed iteratively for $i\geq 2$ and all nodes $v$ in $V$ as
\begin{equation*}\label{eq:morgan}
	\mathsf{ec}^{(i)}(v) = \sum_{u \in N(v)} \mathsf{ec}^{(i-1)}(u),
\end{equation*}
until the number of different values no longer increases. For such an iteration $i$, $\mathsf{ec}^{(i-1)}(v)$ is the final extended connectivity of the node $v$. \citet{Razinger1982} and \citet{Figueras1993} independently observed that the extended connectivity values $\mathsf{ec}^{(i)}$ are equivalent to the row (or column) sums of the $i$th power of the adjacency matrix, which is equal to the number of walks of length $i$ starting at the individual nodes.

The general idea of encoding neighborhoods of increasing radius to make the descriptor more specific resembles the idea of the Weisfeiler--Leman algorithm. However, the connectivity value does not incorporate labels, i.e., atom and bond types, discards the values of $\mathsf{ec}^{(i-1)}(v)$ when computing $\mathsf{ec}^{(i)}(v)$, and loses information due to summation~\citep{Kriege22}.

\paragraph{Circular Fingerprints}
In 1973, \citet{Adamson1973} considered the task of automated classification of chemical structures by representing molecules by (chemical) \new{fingerprints}, i.e., a vectorial representation of a molecule. Today, fingerprints are a standard tool in cheminformatics used to determine molecular similarity, e.g., for classification, clustering, and similarity search in large chemical information systems~\citep{Rogers2010}.
A fingerprint is a vector where each component counts the number of occurrences of certain substructures or merely indicates their presence or absence by a single bit.
Often hashing is used to map substructures to the entries of a fixed-size fingerprint; see~\citet{Daylight2008}.
Fingerprints are typically compared using similarity measures for sets such as the \emph{Tanimoto coefficient}, which satisfies the property of a kernel~\citep{Ralaivola2005}; see~\cref{notation}.
The substructures used may stem from a predefined set obtained either by applying data mining methods, using domain expert knowledge~\citep{Durant2002}, or are enumerated directly from the molecular graph, e.g., all paths up to a given length.

Of particular interest to the present work is the class of \emph{circular fingerprints}, where the substructures are the neighborhoods of each node with increasingly distant nodes added to the neighborhood.
In this sense, this type of fingerprint is conceptually similar to Morgan's algorithm and is occasionally also referred to as \emph{Morgan fingerprint}. However, the key differences are that atom and bond types are encoded, the maximum radius is limited to a typically small value, and all the intermediate results for radii smaller than the maximum are retained~\citep{Rogers2010}.
The earliest of these approaches goes back to so-called \emph{fragments reduced to an environment that is limited} proposed in 1973~\citep{Dubois1973,Dubois1987}. Several variations of the approach have been proposed, e.g., \emph{atom environment fingerprints}~\citep{Bender2004} and \emph{extended connectivity fingerprints}~\citep{Rogers2010}.
These fingerprints are widely used, and implementations are available in open-source and commercial software libraries such as RDKit\footnote{\url{https://www.rdkit.org}} and OpenEye GraphSim TK.\footnote{\url{https://www.eyesopen.com/graphsim-tk}}

\citet{Duv+2015} proposed a neural extension of circular fingerprints introducing learnable parameters for encoding neighborhoods. This work as well as earlier techniques introduced in cheminformatics, e.g.,~\citet{bas+1997,Kir+1995,Mer+2005}, represent early instances of GNNs.

\subsubsection{Graph Neural Networks}
Recently, graph neural networks (GNNs) or message-passing neural networks~\citep{Gil+2017,Sca+2009} (re-)emerged as the most popular machine learning method for graph-structured input.\footnote{In the following, we use the term GNN and message-passing neural network interchangeably; see also~\cref{sec:gnn}} Intuitively, GNNs can be viewed as a differentiable variant of the $1$-WL where colors are replaced with real-valued features and a neural network is used for neighborhood feature aggregation.  By deploying a trainable neural network to aggregate information in local node neighborhoods, GNNs can be trained in an end-to-end fashion together with the classification or regression algorithm's parameters, possibly allowing for greater adaptability and better generalization than the kernel counterpart of the classical $1$-WL algorithm, see~\cref{sec:gnn} for details.

Notable instances of this architecture include~\citet{Duv+2015,Ham+2017,Vel+2018}, which can be subsumed under the message passing framework introduced in~\citet{Gil+2017}. In parallel, approaches based on spectral information were introduced in, e.g.,~\citet{,Bru+2014,Defferrard2016,Gam+2019,Kip+2017,Lev+2019,Mon+2017}. All of the above descend from early work in~\citet{bas+1997,Kir+1995,Mer+2005,mic+2009,mic+2005,Sca+2009,Spe+1997}. Aligned with the field's recent rise in popularity, there exists a plethora of surveys on recent advances in GNN techniques; some of the most recent ones include~\citet{Cha+2020,Wu+2018,Zho+2018}.

\subsubsection{Equivariant Neural Networks}
Input symmetries are frequently incorporated into learning models to construct efficient models. A prominent example is the translation invariance encoded by Convolutional Neural Networks (CNNs), particularly useful for image recognition tasks~\citep{Lec+2015}.  In the last few years, incorporating other types of symmetries in neural networks~\citep{ravanbakhsh2017equivariance,wood1996representation}, e.g., a set structure where the output is invariant to the order of the input~\citep{Zah+2017}, became an important research direction. As with CNNs, the main idea is to construct neural networks as a composition of several (simple) equivariant building blocks, i.e., layers respecting the symmetry; see~\cref{sec:equiv} for details. These networks were shown to reduce the number of free parameters and to improve efficiency and generalization.

One important research direction that follows this line of work is devising equivariant networks for learning on graphs, where, for most tasks, the specific order of nodes does not matter~\citep{albooyeh2019incidence,keriven2019,kondor2018covariant,Mar+2019b,Mar+2019,Mar+2019c,yaronlipman2020global,rav+2020}. In~\cref{sec:equiv}, we discuss these models thoroughly and show that their expressive power is closely related to the Weisfeiler--Leman algorithm.

\subsection{Structure of the Document}
In~\cref{notation}, we fix notation and introduce basic concepts used throughout the present work. \Cref{wl} introduces the $1$-WL, and its generalization, the $k$-WL, and gives an overview of its theoretical properties. In the next section,~\cref{sec:nonneural}, we survey non-neural machine learning approaches leveraging the Weisfeiler--Leman algorithm, focusing on supervised graph classification. \Cref{connect} introduces GNNs and their connection to the $1$-WL and investigates neural architectures beyond $1$-WL's expressive power. Subsequently,~\cref{sec:equiv} describes the recent progress in designing equivariant (higher-order) graph networks and their connection to the Weisfeiler--Leman hierarchy.  Further,~\cref{sec:applications} outlines applications of  Weisfeiler--Leman-based graph embeddings. \Cref{sec:challenges} outlines open challenges and sketches future research directions. Finally, the last section acts as a conclusion.

\section{Preliminaries}\label{notation}

As usual, let $[n] = \{ 1, \dotsc, n \} \subset \NN$ for $n \geq 1$, and let $\{\!\!\{ \dots\}\!\!\}$ denote a multiset.
A  \new{(undirected) graph} $G$ is a pair $(V,E)$ with a \emph{finite} set of
\new{nodes} $V(G)$ and a set of \new{edges} $E(G) \subseteq \{ \{u,v\}
	\subseteq V \mid u \neq v \}$.
For notational convenience, we usually denote an edge $\{u,v\}$ in $E(G)$ by $(u,v)$ or $(v,u)$, and set $n = |V(G)|$ and $m = |E(G)|$. In case of a \new{directed graph}, the set of \new{edges} $E(G) \subseteq \{ (u,v)
	\subseteq V^2 \mid u \neq v \}$, i.e., $E(G)$ might not be symmetric and the edges $(u,v)$ and $(v,u)$ are considered distinct. A \new{labeled graph} $G$ is a triple
$(V,E,l)$ with a label function $l \colon V(G) \cup E(G) \to \Sigma$,
where $\Sigma$ is a subset of the natural numbers. Then $l(w)$ is a
\new{label} of $w$ for $w$ in $V(G) \cup E(G)$. An \new{attributed graph} $G$ is a triple
$(V,E,a)$ with an attribute function $a \colon V(G) \cup E(G) \to \mathbb{R}^d$ for $d>0$. Then $a(w)$ is an \new{attribute} or \new{continuous label} of a node or edge $w$ in $V(G) \cup E(G)$. We denote the set of all labeled or attributed graphs by $\mathcal{G}$.
The \new{neighborhood} of $v$ in $V(G)$ is denoted by $N(v) = \{ u \in V(G) \mid (v, u) \in E(G) \}$.
Let $S \subseteq V(G)$, then the set $S$ induces a \new{subgraph} $(S,E_S)$ with $E_S = \{ (u,v) \in E(G) \mid (u,v) \in S \times S  \}$. We say that two graphs $G$ and $H$ are \new{isomorphic}, denoted $G \simeq H$, if there exists an edge-preserving bijection (\new{graph isomorphism}) $\varphi \colon V(G) \to V(H)$, i.e., $(u,v)$ is in $E(G)$ if and only if $(\varphi(u),\varphi(v))$ is in $E(H)$. In the case of labeled  graphs, we additionally require that
$l(v) = l(\varphi(v))$ for $v$ in $V(G)$, similarly for edge labels. The \emph{graph isomorphism problem} deals with deciding if two graphs are isomorphic or not. The \new{isomorphism type} $\tau(G)$ of a graph $G$ is the equivalence class induced by the (isomorphism) relation $\simeq$, i.e., $\tau(G) = \{ H \in \mathcal{G} \mid G \simeq H \}$.  A \emph{(graph) automorphism} is an isomorphism from a graph to itself, i.e,  $\varphi \colon V(G) \to V(G)$. A \new{(graph) homomorphism} is a map $\varphi \colon V(G) \to V(H)$ where $((\varphi(u),\varphi(v))$ in $E(H)$ if $(u,v)$ in $E(G)$. Note that $\varphi(v) = \varphi(w)$ for two distinct nodes $v$ and $w$ in $V(G)$ is permitted. Hence, as opposed to isomorphisms, homomorphisms need not to preserve non-edges, i.e., $(\varphi(u),\varphi(v))$ in $E(H)$ does not imply $(u,v)\in E(G)$.

\paragraph{Permutation-invariance and -equivariance} Let $n > 0$, then $S_n$ denotes the set of permutations of $[n]$, i.e., the set of all bijections from $[n]$ to itself. Further, let $V(G) = [n]$, then for $\sigma$ in $S_n$, $G_{\sigma} = \sigma \cdot G$ where $V(G_\sigma) = \{ {\sigma(1)}, \dots, {\sigma(n)} \}$ and $E(G_\sigma) = \{ ({\sigma(i)},{\sigma(j)}) \mid (v_i, v_j) \in E(G)  \}$. That is, applying the permutation $\sigma$  reorders the nodes. Hence, for two isomorphic graphs $G$ and $H$, i.e., $G \simeq H$, there exists $\sigma$ in $S_n$ such that $\sigma \cdot G = H$.

Assuming that all graphs have $n$ nodes, a function $f \colon \mathcal{G} \to \bbR$ is \new{invariant} if $f(G) = f(\sigma \cdot G)$ for all graphs $G$ in $\mathcal{G}$ and all permutations $\sigma$ in $S_n$.
More generally, given a set $\mathcal{X}$ on which $S_n$ acts, a function $f \colon \mathcal{G} \to \mathcal{X}$ is \new{equivariant} if $f(\sigma \cdot G) = \sigma \cdot f(G)$. In this paper, we mainly consider $\mathcal{X}=\bbR^n$ as a representation for node features. $S_n$ acts on this space by permuting the entries of the vector, i.e., $\sigma \cdot (x_1,\ldots,x_n) = (x_{\sigma^{-1}(1)},\ldots,x_{\sigma^{-1}(n)})$. Other options for $\mathcal{X}$ are discussed in Section \ref{sec:equiv}.

\paragraph{Kernels} A \emph{kernel} on a non-empty set $\mathcal{X}$ is a positive semi-definite, symmetric function
$k \colon \mathcal{X} \times \mathcal{X} \to \mathbb{R}$.
Equivalently, a function $k$ is a kernel if there is a \emph{feature map}
$\phi \colon \mathcal{X} \to \mathcal{H}$ to a Hilbert space $\mathcal{H}$ with an inner product
$\langle \cdot, \cdot \rangle$, such that
$k(x,y) = \langle \phi(x),\phi(y) \rangle$ for all $x$ and $y$ in $\mathcal{X}$.
Then a positive semi-definite, symmetric function $\mathcal{G} \times \mathcal{G} \to \mathbb{R}$ is a \emph{graph kernel}. Given two vectors $x$ and $y$ in $\mathbb{R}^d$, the \new{linear kernel} is defined as $k(x,y) = x^{\mathsf{T}} y$. Given a finite set $X = \{ x_1, \dots, x_n \} \subseteq \mathcal{X}$  and a kernel $k \colon \mathcal{X}\times\mathcal{X} \to \mathbb{R}$, the \new{Gram matrix} $M$ in $\mathbb{R}^{n \times n}$ contains the kernel values for each pair of elements of the set $X$, i.e., $M_{ij} = k(x_i, x_j)$; see, e.g.,~\citet{Moh+2012}, for details.

\section{The Weisfeiler--Leman Method}\label{wl}

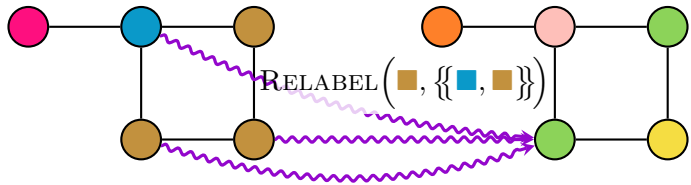
\begin{figure}[t]
	\begin{center}
		\input{figures/wl.tex}
	\end{center}
	\caption{Illustration of $1$-WL's relabeling procedure. The brown, lower-left node gets updated based on the colors of its neighbors.}\label{fig:wlrelabel}
\end{figure}

\begin{figure}[t]
	\begin{subfigure}[b]{0.4\textwidth}
		\centering
		\resizebox{!}{1.5cm}{\input{figures/example1.tex}}
		\caption{Example 1}
	\end{subfigure}%
	\hfill
	\begin{subfigure}[b]{0.6\textwidth}
		\centering
		\resizebox{!}{1.5cm}{\input{figures/example2.tex}}
		\caption{Example 2}
	\end{subfigure}%
	\caption{Examples of two graphs (denoted by \textcolor{pink}{$\blacksquare$} and \textcolor{green}{$\blacksquare$}) that cannot be distinguished by the $1$-WL.\label{fig:wlcounter}}
\end{figure}
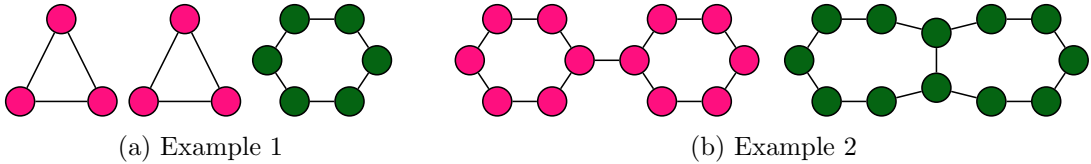

As mentioned in~\cref{sec:intro}, the $1$-WL or color refinement is a simple heuristic for the graph isomorphism problem, originally proposed in~\citet{Wei+1968}.\footnote{Strictly speaking, the $1$-WL and color refinement are two different algorithms. That is, the $1$-WL considers neighbors and non-neighbors to update the coloring, resulting in a slightly higher expressive power when distinguishing nodes in a given graph; see~\citet{gro21} for details. In the case of graph classification, both algorithms have the same expressive power. For brevity, we consider both algorithms to be equivalent.}
Intuitively, the
algorithm tries to determine if two graphs are non-isomorphic by
iteratively coloring or labeling nodes. Given an
initial coloring or labeling of the nodes of both
graphs, e.g., their degree or application-specific information, in
each iteration, two nodes with the same label get different labels if
the number of identically labeled neighbors is not equal. If, after
some iterations, the number of nodes annotated with a specific label is
different in both graphs, the algorithm terminates, and we conclude
that the two graphs are not isomorphic. It is easy to see that the
algorithm cannot distinguish all non-isomorphic graphs; see~\cref {fig:wlcounter} and~\citet{Cai+1992}. Nonetheless, it is a powerful heuristic that can successfully test isomorphism for a
broad class of graphs~\citep{Bab+1979}, see~\cref{theory} for an
in-depth discussion on the algorithm's properties.

Formally, let $G = (V,E,l)$ be a labeled graph, in each iteration, $i > 0$, the $1$-WL computes a node coloring $C^1_i \colon V(G) \to \bbN$,
which depends on the coloring of the neighbors. That is, in iteration $i>0$, we set
\begin{equation}\label{eq:wlColoring}
	C^1_i(v) = \textsc{Relabel}\Big(\!\big(C^1_{i-1}(v),\oms C^1_{i-1}(u)\!\mid\!u \in\!N(v) \!\cms \big)\! \Big),
\end{equation}
where $\textsc{Relabel}$ injectively maps the above pair to a unique natural number, which has not been used in previous iterations. Viewed differently, $C^1_{i-1}$ induces a partitioning of a graph's node set, which is further refined by $C^1_{i}$. In iteration $0$, the coloring $C^1_{0} = l$ or a constant value if no labeling is provided.

That is, in each iteration, the algorithm computes a new color for a node based on the colors of its neighbors; see~\cref{fig:wlrelabel} for an illustration. Hence, after $k$ iterations the color of a node $v$ captures \emph{some} structure of its
$k$-hop neighborhood, i.e., the subgraph induced by all nodes reachable by walks of length at most $k$.

To test if two graphs $G$ and $H$ are non-isomorphic, we run the above algorithm in ``parallel'' on both graphs. If the two graphs have a different number of nodes colored $c$ in $\bbN$ at some iteration, the $1$-WL concludes that the graphs are not isomorphic. Moreover, if the number of colors between two iterations, $i$ and $(i+1)$, does not change, i.e., the cardinalities of the images of $C^1_{i}$ and $C^1_{i+1}$ are equal or, equivalently,
\begin{equation*}
	C^1_{i}(v) = C^1_{i}(w) \iff C^1_{i+1}(v) = C^1_{i+1}(w),
\end{equation*}
for all nodes $v$ and $w$ in $V(G)$, the algorithm terminates. For such $i$, we define the \new{stable coloring}
$C^1_{\infty}(v) = C^1_i(v)$ for $v$ in $V(G)$. The stable coloring is reached after at most $\max \{ |V(G)|,|V(H)| \}$ iterations~\citep{Gro2017}; see~\cref{theory} for further bounds on the algorithm's running time.

\subsection{The \emph{k}-dimensional Weisfeiler--Leman Algorithm}
\label{ss:k_dim_WL_algorithm}

Due to the shortcomings of the $1$-WL  or color refinement in distinguishing non-isomorphic
graphs, several researchers~\citep{Bab1979,Bab2016,Imm+1990},
devised a more powerful generalization of the former, today known
as the \new{$k$-dimensional Weisfeiler--Leman algorithm}.\footnote{In~\citet{Bab2016}, László Babai mentions that he first introduced the algorithm in 1979 together with Rudolf Mathon from the University of Toronto.} In the literature, there exist two variants of the algorithm, which differ slightly in the way they aggregate information. The variant we describe below is often denoted \new{folklore} $k$-dimensional Weisfeiler--Leman algorithm ($k$-FWL) in the machine learning literature, e.g., see~\cite{Mar+2019,Mor+2019}. We follow this convention to be aligned with papers in the machine learning literature. We also define the other variant, named oblivious $k$-WL ($k$-OWL), see~\cref{ss:obvkwl}.

Intuitively, to surpass the limitations of the $1$-WL, the $k$-FWL colors subgraphs instead of a single node. More precisely, given a graph $G$, it colors tuples from $V(G)^k$ for $k \geq 1$ instead of nodes. By defining a neighborhood between these tuples, we can define a coloring  similar to the $1$-WL. Formally, let $G$ be a graph, and let
$k \geq 2$. Moreover, let $\vec{v}$ be a tuple in $V(G)^k$,
then $G[\vec{v}]$ is the subgraph induced by the components of $\vec{v}$, where the nodes are labeled with integers from $\{ 1, \dots, k \}$ corresponding to indices of $\vec{v}$. In each iteration $i \geq 0$, the algorithm, similarly to the $1$-WL, computes a
\new{coloring} $C^k_i \colon V(G)^k \to \bbN$. In the first iteration ($i=0$), two tuples $\vec{v}$ and $\vec{w}$ in $V(G)^k$   get the same
color if the map $v_i \mapsto w_i$ induces an isomorphism between $G[\vec{v}]$ and $G[\vec{w}]$.  Now, for $i > 0$, $C^k_{i+1}$ is defined
by
\begin{equation}\label{ci}
	C^k_{i+1}(\vec{v}) = \textsc{Relabel}\Big(\!(C^k_{i}(\vec{v}), M_i(\vec{v}))\!\Big),
\end{equation}
where the multiset
\begin{equation}\label{mi}
	M_i(\vec{v}) = \{\!\! \{    (C^k_{i}(\phi_1(\vec{v},w)), \dots,  C^k_{i}(\phi_k(\vec{v},w)))   \mid w \in V(G)   \}\!\!\}
\end{equation}
and
\begin{equation*}
	\phi_j(\vec{v},w) = (v_1, \dots, v_{j-1}, w, v_{j+1}, \dots, v_k).
\end{equation*}
That is, $\phi_j(\vec{v},w)$ replaces the $j$-th component of the tuple $\vec{v}$ with the node $w$. Hence, two tuples are \new{adjacent} or \new{$j$-neighbors} (with respect to a node $w$) if they are different in the $j$th component (or equal, in the case of self-loops). Again, we run the algorithm until convergence, i.e.,
\begin{equation*}
	C^k_{i}(\vec{v}) = C^k_{i}(\vec{w}) \iff C^k_{i+1}(\vec{v}) = C^k_{i+1}(\vec{w}),
\end{equation*}
for all $\vec{v}$ and $\vec{w}$ in $V(G)^k$ holds, and call the partition of $V(G)^k$
induced by $C^k_i$ the stable partition. For such $i$, we define
$C^k_{\infty}(\vec{v}) = C^k_i(\vec{v})$ for $\vec{v}$ in $V(G)^k$. Hence, two tuples $\vec{v}$ and $\vec{w}$ with the same color in iteration $(t-1)$ get different colors in iteration $t$ if there exists $j$ in $[k]$ such that the number of $j$-neighbors of $\vec{v}$ and $\vec{w}$, respectively, colored with a certain color is different. The algorithm then proceeds analogously to the $1$-WL.

By increasing $k$, the algorithm gets more powerful in distinguishing non-isomorphic graphs, i.e., for each $k\geq 1$, there are non-isomorphic graphs distinguished by the ($k+1$)-WL but not by the $k$-WL~\citep{Cai+1992}. See~\cref{theory}
for a thorough discussion of the algorithm's properties and limitations.
\begin{figure}[t]
	\centering
	\input{figures/kwl.tex}
	\caption{Illustration of $k$-FWL's neighborhood definition for $k=3$. The $3$-tuple $(w,b,c)$ is a $1$-neighbor of the $3$-tuple $(a,b,c)$, while $(a,v,c)$ is a $2$-neighbor of the $3$-tuple $(a,b,c)$.}\label{kwl}
\end{figure}
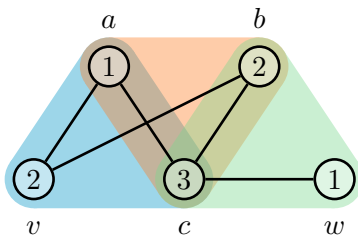

\subsection{Oblivious \emph{k}-WL}\label{ss:obvkwl}

In the literature, e.g.,~\cite{Gro+2000}, there exists a variation of~\cref{mi} that leads to a slightly less powerful algorithm using the coloring $C^{k,*}_i \colon V(G)^k \to \mathbb{N}$. For $i > 0$, $C^{k,*}_{i+1}$ is defined
by
\begin{equation}\label{cio}
	C^{k,*}_{i+1}(\vec{v}) = \textsc{Relabel}\Big(\!(C^{k,*}_{i}(\vec{v}), M^*_i(\vec{v}))\!\Big),
\end{equation}
where $M_i(\vec{v})$ in~\cref{ci} is replaced by
\begin{equation}\label{mio}
	M^*_i(\vec{v}) =   \big( \{\!\! \{  C^{k,*}_i(\phi_1(\vec{v},w)) \mid w \in V(G) \} \!\!\}, \dots, \{\!\! \{  C^{k,*}_i(\phi_k(\vec{v},w)) \mid w \in V(G) \} \!\!\} \big).
\end{equation}
Following~\cite{gro21}, we call the resulting algorithm \new{oblivious} $k$-WL ($k$-OWL).

It holds that the $1$-OWL and $2$-OWL have the same expressive power and that the $(k+1)$-OWL has the same expressive power as the $k$-FWL for $k \geq 2$~\citep{gro21}. The reason the $k$-OWL has a lower expressive power than the $k$-FWL is due to the different way they aggregate colors. That is, the $k$-FWL, see~\cref{mi}, groups colors of $k$-tuples according to the replaced node. For example, by that, the $k$-FWL is able to reconstruct if there is an edge between the two exchanged nodes; see~\citet{gro21} for details.

\subsection{Theoretical Properties}\label{theory}

The Weisfeiler--Leman algorithm constitutes one of the earliest
approaches to isomorphism testing~\citep{Wei+1968,Wei+1976}. The
$1$-dimensional version is an essential building block of the
individualization-refinement approach to graph isomorphism testing
\citep{Mck1981}, forming the basis of almost all practical
graph isomorphism solvers. Higher-dimensional versions have been
heavily investigated by the theory community over the last few
decades, see, e.g., \citet{Cai+1992,Gro2017,Kie2020b,Ott1997}.
For logarithmic $k$, the $k$-dimensional WL algorithm is an essential
building block of Babai's isomorphism algorithm~\citep{Bab2016} running in quasipolynomial time, i.e., its running time is in $2^{\mathcal{O}(\log^c n)}$ for a constant $c > 0$. In the following, we overview the Weisfeiler--Leman algorithm's theoretical properties, stressing relevance for machine learning with graphs when possible.

\paragraph{Expressive Power}
We say that the $k$-FWL or $k$-OWL  \emph{distinguishes} two graphs $G$ and $H$ if their
color histograms differ, i.e., there is some color $c$ in the
image of $C^k_\infty$ such that $G$ and $H$ have different numbers of
node tuples of color $c$. Furthermore, $k$-FWL or $k$-OWL \emph{identifies} a
graph $G$ if it distinguishes $G$ from all graphs not isomorphic to
$G$.

\begin{figure}[t]
	\centering
	\resizebox{!}{4cm}{\input{figures/example3.tex}}
	\caption{Two non-isomorphic strongly regular graphs (denoted by \textcolor{pink}{$\blacksquare$} and \textcolor{green}{$\blacksquare$}) with parameters $(16,6,2,2)$ that cannot be distinguished by $2$-FWL: the line graph of $K_{4,4}$ (left) and the Shrikhande graph (right). Figure adapted from~\citet{Gro+21b}. \label{fig:wl2counter}}
\end{figure}
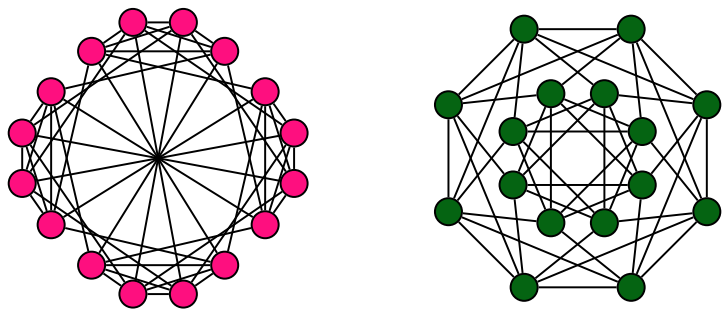

As previously mentioned, \cref{fig:wlcounter} shows a pair of simple, non-isomorphic graphs that
are not distinguished by the $1$-WL. Still, it is likely that the
$1$-WL will distinguish any two random graphs. It can be shown
that the $1$-WL almost surely identifies all graphs. That is, the
probability that the $1$-WL identifies a graph chosen uniformly at random
from the class of all $n$-node graphs goes to $1$ as $n$ goes to
infinity. The above result follows from an old result due to~\citet{Bab+1980}
stating that with probability greater than $1-\sqrt[7]{\nicefrac{1}{n}}$, in a random $n$-node graph, all nodes get different colors after just two iterations of running the $1$-WL. The result
was subsequently refined and extended;
see~\citet{Bab+1979,Cza+2009,Kar+1979,Lip+1979}. While the $1$-WL cannot
distinguish any two regular graphs with the same number of nodes
and degree, \citet{Bol82} showed that the $2$-FWL identifies almost all
$d$-regular graphs for every degree $d$. However,
the $2$-FWL cannot distinguish any two strongly regular graphs with the
same parameters, see, e.g.,~\citet{Gro+21b}, and~\cref{fig:wl2counter}.
For $k\geq 3$, it is much harder to find non-isomorphic graphs that are
not distinguished by the $k$-FWL, resulting in the seminal paper by~\citet{Cai+1992}. For every $k$,
they constructed non-isomorphic graphs $G_k$ and $H_k$, with the number of nodes in $\cO(k)$, that
are not distinguished by the $k$-FWL. These graphs can be
distinguished by the
$(k+1)$-FWL. Hence with increasing dimension, the expressive power of the Weisfeiler--Leman
algorithm increases. This hierarchy of more powerful algorithms was
later leveraged to devise more powerful graph neural networks,
see~\cref{connect,sec:equiv}.

While the construction outlined in~\citet{Cai+1992} shows the limitations of the Weisfeiler--Leman algorithm,
the algorithm is still powerful, in combination with the structural restrictions of the graphs. The \emph{WL dimension} of
a graph $G$ is the least $k$ such that $k$-FWL identifies $G$.
Clearly, every $n$-node graph is identified by $(n-1)$-FWL and thus
has WL-dimension at most $n-1$. In a far-reaching result,~\citet{Gro2012,Gro2017} proved that for every $h > 0$ there is a $k > 0$
such that all graphs, excluding some $h$-node graph
as a minor, have WL-dimension at most $k$. Here a graph $H$ is a \emph{minor} of a graph $G$ if $H$ is isomorphic to a graph obtained from $G$ by deleting
nodes or edges and by contracting edges. Since planar graphs
exclude the complete 5-node graph $K_5$ as a minor, planar graphs have a bounded WL dimension. Similarly, the theorem shows
that graphs of bounded genus or bounded treewidth and also more
esoteric topologically constrained graphs, for example, graphs
that can be embedded into 3-space in such a way that no cycle is
knotted~\citep{Rob+1993}, have bounded WL dimension. Other graphs known to
have bounded WL dimensions are interval graphs~\citep{Evd+2000} and
graphs of bounded rank width~\citep{Gro+2019a}. For some of these
classes, explicit bounds on the WL dimension are known. Most notably,
planar graphs have WL dimension at most $3$~\citep{Kie+2019}. This result has relevance for many applications that involve planar
graphs. For example,  a large portion of molecules is known to be
planar~\citep{Horvath2010a,Yamaguchi2003a}.
Moreover,~\citet{Kie+2015,Arv+2015} gave a complete characterization of the
graphs of WL dimension $1$. See~\citet{Kie2020a,Kie2020b} for thorough overviews of the algorithm's
expressive power.

\paragraph{Complexity}
While a naive implementation of the $1$-WL requires (at least) quadratic
time $\cO(nm)$, where $n$ is the number of nodes and $m$ the number
of edges of the input graph,~\citet{Car+82}
proved that the stable coloring $C^1_\infty$ can be computed in almost
linear time $O(n+m\log n)$; also see \citet{Pai+87}. \citet{Ber+2017a} proved that this is optimal within a large
class of natural partitioning algorithms that includes all known
algorithms for $1$-WL. \citet{Imm+1990} generalized
the almost-linear $1$-WL algorithm to the $k$-FWL and proved that the stable
coloring $C^k_{\infty}$ can be computed in time
$O(k^2 n^{k+1}\log n)$. For every fixed $k\geq 1$, the problem of
deciding whether two graphs are distinguished by $k$-FWL is
\textsf{PTIME}-complete under logspace reductions~\citep{Gro1999}. Hence, it is
unlikely that there are fast parallel algorithms computing the stable
coloring.

Related to these complexity-theoretic results is the question of how many iterations the $k$-FWL needs to reach the stable coloring. A trivial
upper bound is $n^k-1$ because, in each iteration, the number of
colors increases, and a partition of a set of size $n^k$ has at
most $n^k$ classes. \citet{Kie+2020} devised
several infinite classes of graphs where $1$-WL needs the maximum number
of $n-1$ iterations. Quite surprisingly,~\citet{Lic+2019} proved an upper bound of $O(n\log n)$ on the
number of iterations of $2$-FWL, a subquadratic upper bound was already
known from \citet{Kie+2016}. No non-trivial upper bound is known for the
$k$-FWL with $k\ge 3$, and the best known lower bound for all $k$ is
linear \citep{Fur01}.

\paragraph{Connections With Other Areas}

A particularly nice feature of the Weisfeiler--Leman algorithm is that it has several
characterizations in terms of seemingly unrelated concepts from logic,
algebra, and combinatorics. Here, the logical
characterization turned out to be instrumental in proving
several of the expressive power results mentioned above. Specifically, \citet{Cai+1992} showed that two graphs are
indistinguishable by the $k$-FWL if and only if they satisfy the same
sentences of the logic $\textsf{C}^{k+1}$, the $(k+1)$-variable fragment of
first-order logic extended by counting quantifiers.

\citet{Tin1986,Tin1991} derived an equivalence between $1$-WL's
inability to distinguish two non-isomorphic graphs and a system of
linear equations having a real solution. By considering the relaxation
$L$ of an integer linear program for the graph isomorphism problem,
he showed that two non-isomorphic graphs cannot be distinguished by
the $1$-WL if and only if $L$ has a real solution, also known as
\new{fractional isomorphism}. The authors of \citet{Ats+2013,GroheO15,Mal2014} later
lifted the above equivalence to the $k$-FWL by considering a slight
variation $L^k_{-}$ of the linear program $L^k$ of the $k$th level of
the Sherali-Adams hierarchy for the linear program $L$. For $k \geq
	2$, they showed that two non-isomorphic graphs cannot be distinguished
by the $k$-FWL if and only if the $L^k_{-}$ has a real
solution. Similar results were obtained for systems of polynomial
equations~\citep{Ber+2017b}, algebraic proof
systems~\citep{Ber+2015,Gra+19}, semidefinite
programming~\citep{Ats+18,Odo+14}, and non-signaling
quantum isomorphisms~\citep{Ats+2019}. Finally, \citet{Ker+2014} pointed out a close relationship between the $1$-WL and
the Franke-Wolfe algorithm for convex optimization.

\citet{Dvo10}, see also~\citet{Del+2018}, showed a connection between
$k$-FWL's expressive power and homomorphism counts. Given two graphs $F$ and
$G$, $\textsf{hom}(F,G)$ denotes the number of (graph) homomorphisms between
the graphs $F$ and $G$. Given a set of graphs $\mathcal{F}$, the
\new{homormorphism number vector}
$\overrightarrow{\textsf{hom}}(\mathcal{F}, G) =
	(\textsf{hom}(F,G))_{F \in \mathcal{F}}$ contains the number of
homomorphisms between any graph in $\mathcal{F}$ and $G$. \citet{Dvo10} showed that the $k$-FWL does not distinguish a pair of non-isomorphic graphs if and only if their homomorphism number vectors are equal for the set of graphs with treewidth of at most $k$.

\section{Non-neural Methods for Machine Learning Based on the Weisfeiler--Leman Algorithm}
\label{sec:nonneural}
In the following, we review applications of the Weisfeiler--Leman method for machine learning focusing on graph kernels. Hence, this section mainly deals with (supervised) graph-level prediction tasks, e.g., graph classification, where node and edge labels are often absent. Starting from the Weisfeiler--Leman subtree kernel~\citep{She+2011}, we thoroughly survey graph kernels based on the  Weisfeiler--Leman method.

\subsection{Weisfeiler--Lehman Subtree Kernel}

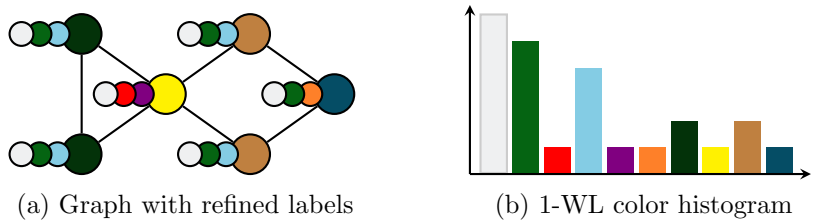
\begin{figure}
	\centering
	\begin{subfigure}[b]{0.4\textwidth}
		\centering
		\input{figures/histogram1.tex}
		\caption{Graph with refined labels}
	\end{subfigure}%
	\begin{subfigure}[b]{0.4\textwidth}
		\centering
		\input{figures/histogram2.tex}
		\caption{$1$-WL color histogram}
	\end{subfigure}%
	\caption{Illustration of the feature vector (color histogram) computed by the Weisfeiler--Lehman subtree kernel. From left to right, smaller circles represent colors from previous iterations. Large circles represent the $1$-WL colors after three iterations. Initially, all nodes are colored gray.}
	\label{fig:wlhist}
\end{figure}

The Weisfeiler--Lehman subtree kernel~\citep{She+2009b} constitutes the earliest approach to leverage the $1$-WL as a graph kernel, inspiring many follow-up works. The primary idea is to compute the $1$-WL for $h \geq 0$ iterations, resulting in a coloring $C^1_i \colon V(G) \to \Sigma_i$ for each iteration $i$, where $\Sigma_i$ is a finite subset of the natural numbers, i.e., $\Sigma_i \subset \bbN$. Notice that the image of the coloring changes in every iteration, depending on the multiset generated by $1$-WL. For $i = 0$, we set $\Sigma_0 = \Sigma$, i.e., the original node label alphabet. In each iteration, we compute a feature vector or color histogram $\phi_i(G)$ in $\bbR^{|\Sigma_i|}$ for each input graph $G$.

Each component $\phi_i(G)_{c}$ counts the number of occurrences of nodes labeled by $c$ in $\Sigma_i$. With the ordering of $\Sigma_i$ being fixed and known beforehand---which is equivalent to knowing the label alphabet~$\Sigma$ in advance---the vector $\phi_i(G)$ can be padded with zeroes if necessary. The overall feature vector $\phi_{\text{WL}}(G)$ is then defined as the concatenation of the feature vectors of all $h$ iterations, i.e.,
\begin{equation}
	\phi_{\text{WL}}(G) = \big[\phi_0(G), \dots, \phi_h(G) \big].
\end{equation}
See~\cref{fig:wlhist} for an illustration of the feature vector $\phi_{\text{WL}}(G)$. We obtain the corresponding kernel for $h$ iterations as
\begin{equation}
	k_{\text{WL}}(G,H) = \langle \phi_{\text{WL}}(G), \phi_{\text{WL}}(H) \rangle,
\end{equation}
where $\langle \cdot, \cdot \rangle$ denotes the standard inner product or linear kernel. Hence, the Weisfeiler--Lehman subtree kernel sums the number of node pairs with the same color over all refinement steps. Note that more powerful kernels may also replace the linear kernel, such as the RBF kernel~(see, e.g.,~\citet{Tog+2019}).

The running time for a single feature vector computation is in $\cO(hm)$ and $\cO(Nhm+N^2hn)$ for the calculation of the Gram matrix for a set of $N$ graphs~\citep{She+2011}, under the assumption that a linear-time perfect hashing function is available for computing the coloring.  Here, $n$ and $m$ denote the maximum number of nodes and edges over all $N$ graphs, respectively. Hence, the algorithm scales well to large graphs and data sets and can be used together with linear SVMs~\citep{Cha+2008} to avoid the quadratic overhead of computing the Gram matrix.

\subsection{Variations of the Weisfeiler--Lehman Subtree Kernel}

The subtree kernel gives rise to many variations, focusing on different aspects of a graph.
\citet{She+2011} describe two variations, which we will briefly discuss.

The first is the \new{Weisfeiler--Lehman edge kernel}, instead of counting the color of nodes, it computes a feature vector $\phi^{\text{E}}_i(G)$, counting edges whose incident nodes have \emph{identical} colors. Two such feature vectors can then be compared using a linear kernel. As for the node-based subtree kernel described above, the overall kernel expression for the edge-based Weisfeiler--Leman kernel is an inner product of feature vectors concatenated over each iteration,
\begin{align*}
	k_{\text{WL}}^{\text{E}}(G, H) =
	\langle \phi^{\text{E}}(G), \phi^{\text{E}}(H) \rangle,
	\shortintertext{where}
	\phi^{\text{E}}(G) = [\phi^{\text{E}}_0(G), \dotsc, \phi^{\text{E}}_h(G)].
\end{align*}
The second variation is obtained similarly to the first one but employs a \emph{shortest-path kernel}~\citep{Bor+2005} in iteration~$i$. This results in a feature vector of the form $\phi^{\text{SP}}(G) = [\phi^{\text{SP}}_0(G), \dotsc \phi^{\text{SP}}_h(G)]$. Each $\phi^{\text{SP}}_i(G)$ consists of triples $(\sigma, \tau, l)$, with $\sigma$ and $\tau$ in $\Sigma_i$ denoting the labels of the start and end node of the shortest path, respectively, and $l$ denoting its length, which can either be an edge count or incorporate additional edge weights of the graph.
Again, such a kernel can be expressed as an inner product of concatenated feature vectors,
\begin{equation*}
	k_{\text{WL}}^{\text{SP}}(G, H) =
	\langle \phi^{\text{SP}}(G), \phi^{\text{SP}}(H) \rangle.
\end{equation*}
The advantage of both of these variations is their flexibility---more complicated kernels can be easily accommodated, making it possible to capture additional information on the edge labels of a graph.

\subsection{Matching-based Kernels}\label{sec:Matching-based kernels}

The Weisfeiler--Lehman subtree kernel sums the number of node pairs with the same color over all refinement steps. Other approaches to graph similarity match node pairs colored by the Weisfeiler--Leman method and obtain a graph kernel from an optimal assignment~\citep{Kriege2016} or the Wasserstein distance~\citep{Tog+2019}, which we overview below.

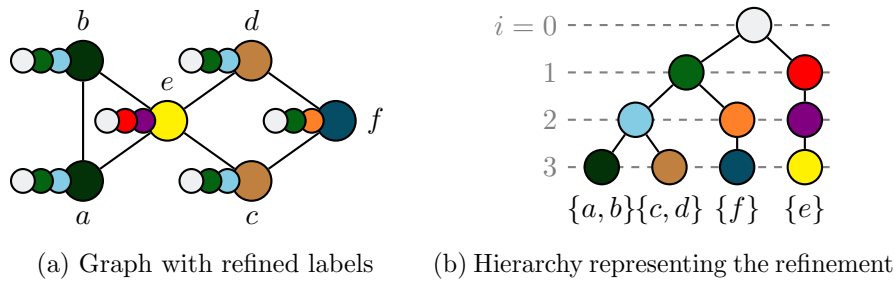
\begin{figure}\centering
	\begin{subfigure}[b]{0.4\textwidth}
		\centering
		\input{figures/assignment_wl.tex}
		\caption{Graph with refined labels}
	\end{subfigure}%
	\begin{subfigure}[b]{0.4\textwidth}
		\centering
		\input{figures/assignment_wl_hierarchy.tex}
		\caption{Hierarchy representing the refinement}
	\end{subfigure}%
	\caption{Hierarchical node partitioning from the 1-WL. Smaller circles represent colors from previous iterations. Large circles represent the $1$-WL colors after three iterations. Initially, all nodes are colored gray.}
	\label{fig:assignment}
\end{figure}
\paragraph{Kernel Based on Optimal Assignments}
Given two sets $A$ and $B$ with $|A|=|B|=n$ and a similarity matrix $S$ in $\bbR^{n\times n}$, where $S_{ij}$ is the similarity of $A_i$ and $B_j$ in $A$ and $B$, respectively, the \emph{(linear) assignment problem} aims to find
\begin{equation}\label{eq:lap}
	\operatorname{LAP}(A,B) = \max_{P\in\mathcal{P}_n} \langle P, S\rangle \quad\text{with}\quad \mathcal{P}_n = \left\{P  \in \{0,1\}^{n\times n} : P \vec{1} = \vec{1}, P^\top \vec{1} = \vec{1} \right\},
\end{equation}
where $\mathcal{P}_n$ is the set of $n \times n$ permutation matrices, $\vec{1}$ is a vector of ``ones,'' and $\langle \cdot, \cdot\rangle$ is the Frobenius inner product, i.e., the element-wise product of two matrices.

Hence, we can compare graphs by computing an optimal assignment between their nodes according to a similarity function defined on their nodes, augmenting the smaller graph with dummy nodes if necessary.
The first graph kernel based on this idea was proposed by~\citet{Fro+2005}. The similarities on the nodes are determined by arbitrary kernels taking the node attributes and their neighborhood into account.
However, in this case, \cref{eq:lap} not always yields a positive semidefinite kernel~\citep{Vert2008}.
\citet{Kriege2016} showed that when the similarity matrix $S$ is obtained from a specific class of base kernels derived from a hierarchy, the value of the optimal assignment is guaranteed to yield a positive semidefinite kernel. Such base kernels can be obtained from the Weisfeiler--Leman method based on the following observation.
The Weisfeiler--Leman method produces a hierarchy on the nodes of a set of graphs, where the $i$th level
consists of nodes for each color in the refinement step $i+1$ with an artificial root at level $0$.
The parent-child relationships are given by the refinement process, where the root has the initial node labels as children, see \cref{fig:assignment}.
This hierarchy gives rise to the base kernel
\begin{equation}\label{eq:wloa:vsim}
	k(u,v) = \sum_{i=0}^h k_\delta(C^1_i(u),C^1_i(v)), \quad k_\delta(x,y) =
	\begin{cases}
		1, \ x=y \\
		0, \ x\neq y
	\end{cases}
\end{equation}
on the nodes.
The kernel counts the number of iterations required to assign different colors to the nodes and reflects the extent to which the nodes have a structurally similar neighborhood. For example, in \cref{fig:assignment}, we have $k(a,f) = 2$, because the nodes $a$ and $f$ are contained in the same subtree on level $0$ and $1$, but not on the deeper levels.
The optimal assignment kernel with this base kernel is referred to as the \emph{Weisfeiler--Lehman optimal assignment kernel}. It is computed in linear time from the hierarchy of the base kernel and achieves better accuracy results in many classification experiments compared to the Weisfeiler--Lehman subtree kernel.
Moreover, the hierarchy can be endowed with weights, which can be optimized via multiple kernel learning~\citep{Kri2019}.

\paragraph{Kernel Based on Wasserstein Distances}
A related idea to establish an optimal matching is employed by the so-called \emph{Wasserstein distance} (or \emph{earth mover's distance}, \emph{optimal transport distance}).
Let $A$ and $B$ in $\bbR^n_+$ with entries that sum to the same value and $D$ in $\mathbb{R}^{n\times n}_+$ a distance matrix, the \emph{Wasserstein distance}\footnote{Depending on the context, slightly different definitions are used in the literature, often requiring that $A$ and $B$ are probability distributions.}
\begin{equation}\label{eq:wasserstein}
	W(A,B) = \min_{T\in\Gamma(A,B)} \langle T, D\rangle \quad\text{with}\quad \Gamma(A,B) = \left\{T  \in \bbR^{n\times n}_+ : T \vec{1} = A, T^\top \vec{1} = B \right\},
\end{equation}
where $\Gamma(A,B)$ is the set of so-called \emph{transport plans}. Intuitively, an element $D_{ij}$ of the matrix $D$ specifies the cost of moving one unit from $i$ to $j$. Then, the Wasserstein distance is the minimum cost required to transform $A$ into $B$.

\citet{Tog+2019} derived valid kernels from the Wasserstein distance by using a distance on the nodes obtained from the Weisfeiler--Leman method according to
\begin{equation}\label{eq:wwl:vdist}
	d(u,v) = \frac{1}{h+1}\sum_{i=0}^h \rho(C^1_i(u),C^1_i(v)), \quad\rho(x,y) =
	\begin{cases}
		1, \ x\neq y \\
		0, \ x=y.
	\end{cases}
\end{equation}
\Cref{eq:wwl:vdist} can be regarded as a normalized distance associated with the kernel of \cref{eq:wloa:vsim}.
For example, in \cref{fig:assignment}, we have $d(a,f) = \nicefrac{1}{2}$, since the nodes $a$ and $f$ are in different subtrees in two of the four levels.
The Wasserstein distance $W(A,B)$ of~\cref{eq:wasserstein} using~\cref{eq:wwl:vdist} on the nodes is then combined with a distance substitution kernel~\citep{Haasdonk2004}, specifically a variant of the Laplacian kernel. The resulting kernel was shown to be positive semidefinite.
For graphs with continuous attributes, \citet{Tog+2019} proposed an extension of the Weisfeiler--Leman method replacing discrete colors with real-valued vectors. Then, the ground costs of the Wasserstein distance are obtained from the Euclidean distance between these vectors. In this case, it is not guaranteed that the resulting kernel is positive semidefinite. To circumvent this issue, \citet{Tog+2019} proposed using a {K}re{\u \i}n SVM~\citep{Loosli2015}, i.e., an SVM that is capable of handling indefinite kernels.

\subsection{Continuous Attributes}

Due to its origin in graph isomorphism testing, the Weisfeiler--Leman algorithm initially only applies to (discretely) labeled graphs. Hence, it is not clear how to extend the algorithm to graphs with continuous attributes, i.e., graphs whose nodes and edges exhibit high-dimensional feature vectors in some $\mathbb{R}^d$. Over the years, there have been multiple noteworthy approaches to address this problem, two of which we briefly discuss and describe below.

The chronologically first approach is due to \citet{Ors+2015} and describes \emph{graph invariant kernels}. The overarching idea is to extend existing graph kernels so that they are able to capture continuous attributes.
This necessitates the definition of a graph invariant. We will provide an abstract definition first and discuss a more concrete example based on $1$-WL later on. A function $\mathcal{I} \colon \mathcal{G} \to \mathbb{R}$ is a \emph{graph invariant} if it maps isomorphic graphs $G$ and $H$ to the same element, i.e., $\mathcal{I}(G) = \mathcal{I}(H)$ if $G \simeq H$. Similarly, a function $\mathcal{L} \colon V(G) \to \mathbb{R}$ is a \emph{node} invariant if it assigns labels to the nodes of a graph $G$ such that they are preserved under any isomorphism~$\varphi$, i.e.,  $\mathcal{L}(v) = \mathcal{L}(\varphi(v))$ for all $v$ in $V(G)$. Since every node invariant can be phrased as a specific graph invariant, we will subsequently not distinguish between node and graph invariants.
Given a graph invariant, we can define a generic kernel function of the form
\begin{equation*}
	k_{\text{GIK}}(G, H) = \sum_{v \in V(G)} \sum_{v' \in V(H)} \mathrm{w}(v, v') \cdot k_{\text{Attr}}(v, v'),
\end{equation*}
where $\mathrm{w}(v, v')$ denotes a function that assesses the similarity between vertices $v$ and $v'$~(\citet{Ors+2015} suggest using a graph invariant function; as we shall subsequently see, $1$-WL can be employed here), and $k_{\text{Attr}}(v, v')$ denotes a kernel between node attributes. A simple choice for $k_{\text{Attr}}(v, v')$ is an RBF kernel. As for $\mathrm{w}(v, v')$, this function can be realized using, among others, the $1$-WL, by setting
\begin{equation}
	\mathrm{w}(v, v') = \sum_{i=0}^{h} \left|\{v \mid c_l^{(i)}(v) = c_l^{(i)}(v')\}\right|,
\end{equation}
i.e., the number of times two nodes are being assigned the same color during the 1-WL refinement scheme with~$h$ iterations. While only being one specific choice for $\mathrm{w}(v, v')$, this demonstrates the utility of $1$-WL beyond the use of a similarity measure itself. In effect, the $1$-WL can also provide more fundamental insights into the structure of a graph.

The second approach is due to \citet{Mor+2016}. Its fundamental idea is to employ the $1$-WL scheme to assess the similarity of labeled graphs, which are, in turn, obtained by employing a hashing scheme. The hashing scheme transforms continuous node attributes into discrete ones, while the 1-WL scheme facilitates the comparison of such labeled graphs. Formally, given a family $\mathcal{H}$ of hash functions, the \emph{hash graph kernel} takes the form
\begin{equation}
	k_{\text{HGK}}(G, H) = \frac{1}{J}\sum_{j=1}^{J} k_{\text{WL}}(h_j(G), h_j(H)),
\end{equation}
where $h_j\colon\mathbb{R}^d \to \mathbb{N}$ refers to a hash function from $\mathcal{H}$. This representation once again demonstrates the versatility of the $1$-WL framework. Multiple hash functions are used in the previous equation to ensure that continuous attributes are represented sufficiently. Originally, the authors propose to use locality-sensitive hashing schemes~\citep{Dat+2004}, but other choices are also possible. The running time of a hash graph kernel evaluation can be upper-bounded by the running time of the $1$-WL scheme, i.e., $\mathcal{O}(hm)$, and the complexity of the hashing scheme $\mathcal{O}(t_h)$, leading to an overall complexity of $\mathcal{O}(J(hm+t_h))$. For a fixed number of iterations~$J$ and under the~(reasonable) assumption that the hash function is no more complex than calculating $1$-WL feature vectors, the hash graph kernel complexity is thus asymptotically no higher than the complexity of $1$-WL.

In addition to these principled approaches, other works also provide variants of the $1$-WL scheme to target continuous attributes. The work by~\citet{Tog+2019}, for instance, which we discussed in~\cref{sec:Matching-based kernels}, can also be applied to graphs with continuous attributes. Its formulation does not give rise to a positive semidefinite kernel, thus necessitating the use of a special SVM for training~\citep{Loosli2015}. Moreover, due to its reliance on Wasserstein calculations, its complexity is considerably higher with $\mathcal{O}(n^3\log n)$ for evaluating the kernel between two graphs~$G$ and $H$, where $n$ refers to the maximum number of nodes in the two graphs.

\subsection{Kernels Based on the \emph{k}-OWL}\label{kwlk}
\citet{Mor+2017} proposed the first graph kernel based on the $k$-OWL. Essentially, the kernel computation works the same way as in the $1$-dimensional case, i.e., a feature vector is computed for each graph based on color counts. To make the algorithm more scalable, the author resorted to coloring all subgraphs on $k$ nodes instead of all $k$-tuples, resulting in a less powerful algorithm~\citep{Abb+2021}. Moreover, the authors proposed only considering a subset of the original neighborhood to exploit the sparsity of the underlying graph. Formally, let $G$ be a graph, for a given $k \geq 2$, they consider all $k$-element subsets $[V(G)]^k$ over $V(G)$. Let $s =  \{ s_1, \dotsc, s_k \}$ be a $k$-set in $[V(G)]^k$, then they define the global neighborhood of $s$ as
\begin{equation*}\label{eq:localNeighborhood}
	N(s) = \{ t\in [V(G)]^k\mid |s\cap t|=k-1\}\,.
\end{equation*}
That is, two $k$-element subsets are neighbors if they are different in one element. The \emph{local neighborhood}\, $N_L(s)$ consists of all $t$ in $N(s)$ such that $(v,w)$ in $E(G)$ for the unique $v$ in $s\setminus t$ and the unique $w$ in $t\setminus s$. The coloring $[V(G)]^k \to \bbN$ is then defined analogously to~\cref{eq:wlColoring} using the local neighborhood. Intuitively, the global neighborhood of a $k$-element subset $s$ consists of all other $k$-element subsets $s$ such that we can go from $s$ to $t$ by replacing exactly one node. The local neighborhood requires that these replaced nodes are adjacent. For example, in~\cref{kwl}, the subset $\{ a, c, v \}$ is in the local neighborhood of $\{ a, b, c \}$ because the nodes $b$ and $v$ are adjacent.

Further, they offered a sampling-based  algorithm to speed up the kernel computation for large graphs approximating it in constant time, i.e., independent of the number of nodes and edges, with an additive approximation error. Finally, they show empirically that the proposed kernel beats the Weisfeiler--Leman subtree kernel on a subset of tested benchmark data sets.

Similarly to the above work,~\citet{Morris2020b} also proposed graph kernels based on $k$-OWL. Again, for scalability, they only consider a subset of the original neighborhood. However, they consider $k$-tuples and prove that a variant of their method is slightly more powerful than the  $k$-OWL, see~\cref{ss:obvkwl} while taking the original graph's sparsity into account. That is, instead of \cref{mi}, it uses
\begin{equation*}\label{midd}
	\begin{split}
		M^{\delta}_i(\vec{v}) =   \big( \{\!\! \{ C^{k,\delta}_{i}(\phi_1(\vec{v},w)) \mid w \in N(v_1) \} \!\!\}, \dots, \{\!\! \{  C^{k,\delta}_{i}(\phi_k(\vec{v},w)) \mid w \in N(v_k) \}  \!\!\} \big).
	\end{split}
\end{equation*}
Hence, two tuples $\vec{v}$ and $\vec{w}$ are \new{local $i$-neighbors} if the nodes $v_i$ and $w_i$ are adjacent in the underlying graph, effectively exploiting the sparsity of the underlying graph. Consequently, the labeling function is defined by
\begin{equation*}\label{ck}
	C^{k,\delta}_{i+1}(\vec{v}) = \textsc{relabel}(C^{k,\delta}_{i}(\vec{v}), M^{\delta}_i(\vec{v})).
\end{equation*}
This local version is incomparable to the $k$-OWL in terms of distinguishing non-isomorphic graphs. That is, there exist pairs of non-isomorphic graphs that the above local variant can distinguish while the $k$-OWL can not and vice versa. However, the authors devised a variant of the above coloring function, with the same asymptotic running time as the above, that is more powerful than the $k$-OWL in distinguishing non-isomorphic graphs. Empirically, they show that this variant of the $k$-OWL achieves a new state-of-the-art across many standard benchmark data sets~\citep{Mor+2020} while being several orders of magnitude faster than the $k$-OWL.

Finally,~\citet{Mor+2022} introduced a more scalable variant of the above local version by omitting certain $k$-tuples. Concretely, they proposed the local $(k,s)$-WL, which only considers $k$-tuples inducing at most $s$ connected components, and studied its expressive power.

\subsection{Other Kernels Based on the 1-WL}

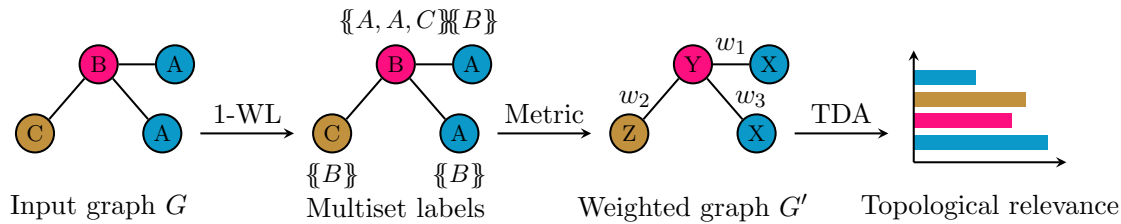
\begin{figure}[tbp]
	\centering
	\resizebox{\linewidth}{!}{%
		\input{figures/p-wl}
	}%
	\caption{%
		A brief overview of the topology-based extension of the $1$-WL scheme, introduced by \citet{Rie+2019}. After obtaining the $1$-WL multiset labels, they are used to define a metric on the input graph~$G$, turning it into a weighted graph~$G'$. This weighted graph is then used to determine the topological relevance of each node, which is subsequently used to reweight the WL feature vector $\phi_\text{WL}$.}
	\label{fig:P-WL}
\end{figure}

The general utility of the $1$-WL scheme made it a natural building block in other algorithms and a central element in others. To guide the subsequent discussion, we briefly expand on the $\mathcal{R}$-convolution framework~\citep{Hau1999}, which to this date underlies most graph kernel approaches either implicitly or explicitly. This framework provides a way to construct kernels to compare structured objects by decomposing them according to a set of agreed-upon substructures, such as shortest paths. Two objects~(e.g., graphs) are then compared by defining a kernel on their respective substructures. Many existing graph kernels can be rephrased as kernels based on the $\mathcal{R}$-convolution framework, see, e.g., \citet{Borg+2020} or \citet{Kriege2020b}, for recent surveys that provide in-depth discussions of this framework.

As an example of an algorithm in which the $1$-WL scheme constitutes a building block, \citet{Yan+2015} employed it in its capacity to enumerate substructures, with the expressed goal to obtain ``smoothed'' variants of existing graph kernels. These are graph kernels built on a less rigid version of the $\mathcal{R}$-convolution framework that supports partial matches between substructures. The smoothed variant of $1$-WL demonstrates superior predictive performance than its ``rigid'' variant but with higher computational costs. In a similar vein, \citet{Yan+2015a} describe how to modify existing graph kernels such that they decompose graphs into their substructures. These substructures are then treated as sentences~(in the natural language processing sense) arising from some vocabulary. This perspective enables the re-weighting of structures based on co-occurrence counts, resulting in a generic kernel formulation
\begin{equation*}
	k_\text{DGK}(G, H) = \phi(G) \mathcal{D} \phi^{\top}(H),
\end{equation*}
where $\phi(\cdot)$ refers to a feature vector representation of a graph kernel and $\mathcal{D}$ denotes a diagonal matrix containing substructure weights. The re-weighted ``deep'' version of $1$-WL also performs slightly better than the unweighted one, but the computational requirements are again substantially higher.

As an example of the second type of approach, where $1$-WL constitutes a critical element, we briefly summarize a method by \citet{Rie+2019}. This paper is motivated by the observation that $1$-WL on its own cannot capture arbitrary topological features, such as cycles, in graphs (see~\cref{theory} or~\citet{Gro+2021} for more details and see~\cref{fig:wlcounter} for a simple example of this). Making use of recent advances in topological data analysis, see \citet{Hen+2021} for a current survey, \citet{Rie+2019} used the ``persistence'', i.e., a type of multi-scale measure for assessing the relevance of topological structures in a graph $G$, to provide weights for the individual dimensions of $1$-WL feature vectors~$\phi_{\text{WL}}(G)$. This amounts to imbuing the label counts with additional information about their topological relevance in terms of connected components and cycles. For instance, if a set of labels often occurs as a part of a pronounced cycle in the graph, its weight will be larger than that of a label that only contributes marginally to the overall topology of a graph.
\Cref{fig:P-WL} illustrates the overall workflow. The $1$-WL is used to generate multiset labels, from which a weighted graph is obtained via a multiset distance metric. Topological features of the graph are then calculated, resulting in a \emph{topological relevance} score for each node or edge.

\begin{figure}[t]
	\begin{center}
		\scalebox{1.3}{\input{figures/gnn.tex}}
	\end{center}
	\caption{Illustration of GNN's neighorbood aggregation.}\label{gnns}
\end{figure}
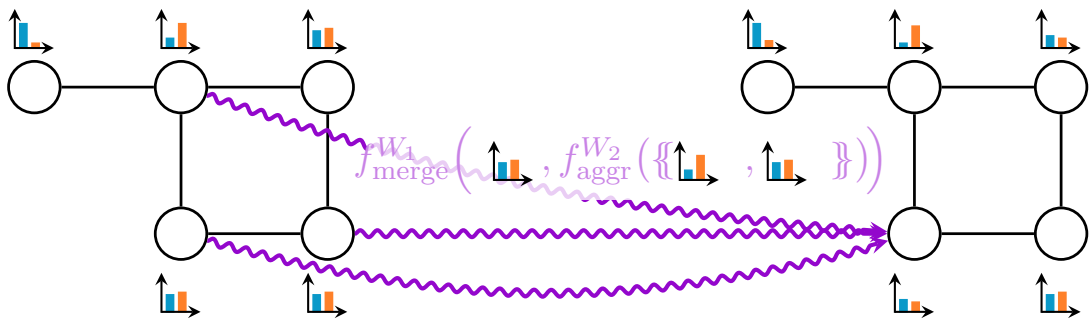

\citet{Rie+2019} empirically demonstrated that the inclusion of cycles can boost the performance of the $1$-WL, particularly for molecular data sets. Moreover, they also proved that it is possible to rephrase the $1$-WL scheme as a specific instance of a general topological relabeling scheme based on graph distances. In essence, the original $1$-WL feature vectors are obtained by using the uniform graph metric, which assigns all edges the same value.
\cite{Zha+2018} devised a pooling method for GNNs, see below, inspired by the $1$-WL histogram construction.

\section{Connections to Graph Neural Networks}
\label{connect}

In the following, we overview the connections between the Weisfeiler--Leman hierarchy, see~\cref{wl}, and neural networks for graphs, specifically GNNs. We introduce GNNs and their connection to the $1$-WL and overview GNN architectures overcoming the limitation of the $1$-WL.

\subsection{GNNs and the 1-WL algorithm}\label{sec:gnn}
Intuitively, GNNs or message-passing neural networks compute a vectorial representation, i.e., a $d$-dimensional vector, representing each node in a graph by aggregating information from neighboring nodes; see~\cref {gnns} for an illustration. Formally, let $G = (V,E,l)$ be a labeled graph with initial node features $f^{(0)} \colon V(G)\rightarrow \RR^{1\times d}$ that are \emph{consistent} with $l$. That is, each node $v$ is annotated with a feature $f^{(0)}(v)$ in $\bbR^{1\times d}$ such that $f^{(0)}(u) = f^{(0)}(v)$ if $l(u) = l(v)$, e.g., a one-hot encoding of the the labels $l(u)$ and $l(v)$. Alternatively, $f^{(0)}(v)$ can be an arbitrary real-valued feature vector or attribute of the node $v$, e.g., physical measurements in the case of chemical molecules.
A GNN architecture consists of a stack of neural network layers, i.e., a composition of parameterized functions. Each layer aggregates local neighborhood information, i.e., the neighbors' features, around each node and then passes this aggregated information on to the next layer.

GNNs are often realized as follows~\citep{Mor+2019}.
In each layer, $t > 0$,  we compute node features
\begin{equation}\label{eq:basicgnn}
	f^{(t)}(v) = \sigma \Big( f^{(t-1)}(v) \cdot  W^{(t)}_1 +\, \sum_{\mathclap{w \in N(v)}}\,\, f^{(t-1)}(w) \cdot W_2^{(t)} \Big)
\end{equation}
in  $\bbR^{1 \times e}$ for $v$, where
$W_1^{(t)}$ and $W_2^{(t)}$ are parameter matrices from $\bbR^{d \times e}$
, and $\sigma$ denotes an entry-wise non-linear function, e.g., a sigmoid or a ReLU function.\footnote{For clarity of presentation, we omit biases.} Following~\citet{Gil+2017,Sca+2009}, one may also replace the sum defined over the neighborhood in the above equation by an arbitrary, differentiable function, and one may substitute the outer sum, e.g., by a column-wise vector concatenation. Thus, in full generality a new feature $f^{(t)}(v)$ is computed as
\begin{equation}\label{eq:gnngeneral}
	f^{W_1}_{\text{merge}}\Big(f^{(t-1)}(v) ,f^{W_2}_{\text{aggr}}\big(\oms f^{(t-1)}(w) \mid  w \in N(v)\cms \big)\!\Big),
\end{equation}
where $f^{W_2}_{\text{aggr}}$ aggregates over the multiset of neighborhood features and $f^{W_1}_{\text{merge}}$ merges the node's representations from step $(t-1)$ with the computed neighborhood features.
Both $f^{W_1}_{\text{aggr}}$ and $f^{W_2}_{\text{merge}}$ may be arbitrary differentiable functions and, by analogy to Equation \ref{eq:basicgnn}, we denote their parameters as $W_1$ and $W_2$, respectively. To adapt the parameters $W_1$ and $W_2$ of Equations \ref{eq:basicgnn}--\ref{eq:gnngeneral}, they are optimized in an end-to-end fashion, usually via a variant of stochastic gradient descent, e.g.,~\citet{Kin+2015}, together with the parameters of a neural network used for classification or regression.

Concurrently with~\citet{Xu+2018b},~\citet{Mor+2019} showed that any GNN's expressive power is upper bounded by the $1$-WL in terms of distinguishing non-isomorphic graphs. That is, given two non-isomorphic graphs, for any choice of functions $f^{W_1}_{\text{merge}}$ and $f^{W_2}_{\text{aggr}}$ and parameters $W_1$ and $W_2$, the GNN is not able to learn node features distinguishing two graphs if the $1$-WL cannot distinguish them. Let $W^{(t)}$ denote the set of weights up to layer $t$. Formally, we can write the above down as follows.

\begin{theorem}[{\citealp{Mor+2019,Xu+2018b}}]\label{thm:refine}
	Let $G=(V,E,l)$ be a labeled graph. Then for all $t\ge 0$ and for all choices of initial colorings $f^{(0)}$ consistent with $l$, and weights $W^{(t)}$,
	\begin{equation*}
		C^1_t(u) = 	C^1_t(v) \text{ implies }     f^{(t)}(u) = f^{(t)}(v),
	\end{equation*}
	for all nodes $u$ and $v$ in $V(G)$.
\end{theorem}

On the positive side,~\citet{Mor+2019} proved that there exists a sequence of parameter matrices ${W}^{(t)}$ such that GNNs have exactly the same expressive power in terms of distinguishing non-isomorphic (sub-)graphs as the $1$-WL algorithm by deriving injective variants of the functions $f^{W_1}_{\text{merge}}$ and $f^{W_2}_{\text{aggr}}$.

This equivalence in expressive power even holds for the simple architecture of~\eqref{eq:basicgnn}, provided one chooses the encoding of the initial labeling $l$ in such a way that different labels are encoded by linearly independent vectors~\citep{Mor+2019}.

\begin{theorem}[{\citealp{Mor+2019}}]\label{equal}
	Let $G = (V,E,l)$ be a labeled graph. Then for all \mbox{$t\geq 0$}, there exists a sequence of weights ${W}^{(t)}$, and a GNN architecture such that
	\begin{equation*}
		C^1_t(u) = 	C^1_t(v) \text{ if and only if }     f^{(t)}(u) = f^{(t)}(v),
	\end{equation*}
	for all nodes $u$ and $v$ in $V(G)$.
\end{theorem}
Similarly,~\citep{Xu+2018b} derived the \new{Graph Isomorphism Network} (GIN) layer and showed that it has the same expressive power as the $1$-WL in terms of distinguishing non-isomorphic graphs. Concretely, the GIN layer updates a feature of node $v$ at layer $t$ as
\begin{equation*}\label{gin}
	{f}^{t}(v) = \text{MLP}\Big((1+ \varepsilon) \cdot  f^{t-1}(v) + \sum_{w \in N(v)} f^{t-1}(w) \Big),
\end{equation*}
where $\text{MLP}$ is a standard multi-layer perceptron, and $\varepsilon$ is a learnable scalar value. See~\citet{gro21} for an in-depth discussion of both approaches. Further,~\citet{Aam+2022} devised an improved analysis using randomization. In summary, we arrive at the following insight: \emph{Any possible graph neural network architecture can be at most as powerful as the $1$-WL in terms of distinguishing non-isomorphic graphs. A GNN architecture has the same expressive power as the $1$-WL if the functions $f^{W_1}_{\text{merge}}$ and $f^{W_2}_{\text{aggr}}$ are injective.}

\citet{barcelo2020} further tightened the relationship between $1$-WL and GNNs by deriving a GNN architecture that has the same expressive power as the logic $\textsf{C}^2$, see~\cref{theory}.
Moreover,~\citet{Gee+2020a} showed a connection between the $1$-WL and the GCN layer introduced in~\citet{Kip+2017}.

\subsection{Neural Architectures Beyond 1-WL's Expressive Power}

In the following, we overview some recent works overcoming the limitations of the $1$-WL.
\paragraph{Higher-order Architectures}

\citet{Mor+2019} proposed the first GNN architecture that overcame the limitations of the $1$-WL. Specifically, they introduced so-called \emph{$k$-GNNs}, which work by learning features over the set of subgraphs on $k$ nodes instead of nodes by defining a notion of the neighborhood between these subgraphs. Formally, let $G$ be a graph, for a given $k$, they consider all $k$-element subsets $[V(G)]^k$ over $V(G)$. Let $s =  \{ s_1, \dotsc, s_k \}$ be a such $k$-element subset, an element in $[V(G)]^k$, then they define the \emph{neighborhood} of $s$ as
\begin{equation*}\label{eq:localNeighborhood_neural}
	N(s) = \{ t\in [V(G)]^k\mid |s\cap t|=k-1\}.
\end{equation*}
That is, two $k$-element subsets are neighbors if they are different in one element. The \emph{local neighborhood}\, $N_L(s)$ consists of all $t$ in $N(s)$ such that $(v,w)$ in $E(G)$ for the unique $v$ in $s\setminus t$ and the unique $w$ in $t\setminus s$. The \emph{global neighborhood} $N_G(s)$ then is defined as $N(s) \setminus N_{\text{L}}(s)$. Hence, the neighborhood definition equals the one of~\cref{kwlk}

Based on this neighborhood definition, one can generalize most GNN layers for node embeddings, e.g., the one from~\cref{eq:basicgnn}, to more powerful subgraph embeddings. Given a graph $G$, in each layer $t$, a $d$-dimensional real-valued feature for a subgraph $s$ can be computed as
\begin{equation}\label{kgnns}
	f^{t}_{k}(s) =  \sigma  \Big(  f^{t-1}_{k}(s) \cdot W_1^{t} + \sum_{u \in N_L (s) \cup N_G (s)} f^{t-1}_{k}(u) \cdot W_2^{t}\Big).
\end{equation}
At initialization, i.e., layer $t=0$, the feature of the $k$-element subset $s$ is set to a one-hot encoding of the (labeled) isomorphism type of the graph $G[s]$ induced by $s$, possibly enhanced by application-specific node and edge features. The authors resort to sum over the local neighborhood in the experiments for better scalability and generalization, showing a significant boost over standard GNNs on a quantum chemistry benchmark data set~\citep{Ram+2014,Wu+2018}.

Moreover, rather than starting at $k$-node subgraphs,~\citet{Mor+2019} also proposed a hierarchical variant of the layer in~\cref{kgnns} that combines the information of the $k$-node subgraph's isomorphism types with learned vectorial representations of $(k-1)$-node subgraphs using a $(k-1)$-GNN. That is, rather than simply using one-hot indicator vectors as initial feature inputs in a $k$-GNN, they proposed a hierarchical variant of $k$-GNN that uses the features learned by a $(k-1)$-dimensional GNN, in addition to the (labeled) isomorphism type, as the initial features, i.e.,
\begin{equation*}
	f^{(0)}_{k}(s) =  \sigma \Big(\big[f^{\text{iso}}(s), \sum_{u \subset s} f^{\,(T_{k-1})}_{k-1}(u) \Big] \cdot W_{k-1} \Big),
\end{equation*}
for some $T_{k-1} > 0$, where $W_{k-1}$ is a matrix of appropriate size, $f^{\text{iso}}$ is a neural network that learns a vectorial representation of the subset $s$ based on a one-hot encoding of the (labeled) isomorphism type of the graph $G[s]$ induced by $s$, and square brackets denote column-wise matrix concatenation. Hence, the features are recursively learned from dimensions $1$ to $k$ in an end-to-end fashion. Further,~\cite{Morris2020b} devised a neural version of the local version of the $k$-OWL, see~\cref{kwlk}, inheriting its expressive power.

\paragraph{Unique Node Identifiers}

\citet{Vig+2020} extended the expressive power of GNNs, by using unique node identifiers, generalizing the message-passing scheme proposed by~\citet{Gil+2017}, see~\Cref{eq:gnngeneral}, by computing and passing matrix features instead of vector features. Formally, given an $n$-node graph, each node $i$ maintains a matrix $U_i$ in $\mathbb{R}^{n \times c}$ for $c > 0$, denoted \emph{local context}, where the $j$-th row contains the node $i$'s vectorial representation of node $j$. At initialization, each local context $U_i$ is set to a one-hot vector in $\mathbb{R}^{n \times 1}$. Now at each layer $l$, similar to the above message-passing framework, the local context is updated as
\begin{equation*}
	U_i^{(l+1)} = u^{(l)}\hspace{-2px}\left(U_i^{(l)},  \tilde{U}_i^{(l)} \right) \in \mathbb{R}^{n \times c_{l+1}} \quad \text{with} \quad \tilde{U}_i^{(l)} = \phi\left(\left\{m^{(l)}({U}_i^{(l)}, {U}_j^{(l)}, {e}_{ij} ) \right\}_{j \in N(i)}\right),
\end{equation*}
where $u^{(l)}, m^{(l)}$, and $\phi$ are update, message, and aggregation functions, respectively, to compute the updated local context, and $e_{ij}$ denotes the edge feature shared by node $i$ and $j$. The authors studied the expressive power of the above architecture, showing that 
it is more powerful than 1-WL, and proposed more scalable alternative variants of the above architecture. Moreover, they derived conditions for equivariance. Finally, promising results on standard benchmark data sets are reported.

To derive more powerful graph representations,~\citet{Mur+2019,Mur+2019b}, inspired by~\citet{Yar+2018}, proposed  \emph{relational pooling}. To increase the expressive power of GNN layers, they averaged over all permutations of a given graph. Formally, let $G$ be a graph, then a representation
\begin{equation}
	f(G) = \frac{1}{|V(G)|!} \sum_{\pi \in \Pi} g(A_{\pi,\pi},[F_\pi, I_{|V|} ])
\end{equation}
is learned, where $\Pi$ denotes all possible permutations of the rows and columns of the adjacency matrix of the graph $G$. Here, $A_{\pi,\pi}$ permutes the rows and columns of the adjacency matrix $A$ according to the permutation $\pi$ in $\Pi$, similarly $F_\pi$ permutes the rows of the feature matrix $F$. Moreover, $g$ is a (possibly permutation-sensitive) function to compute a vectorial representation of the graph $G$, based on  $A_{\pi,\pi}$ and $F_\pi$, $I_{|V|}$ is the $|V| \times |V|$ identity matrix, and $[\cdot, \cdot]$ denotes column-wise matrix concatenation. The authors showed that the above architecture is more powerful in terms of distinguishing non-isomorphic graphs than the $1$-WL, and proposed sampling-based techniques to speed up the computation.  If the underlying model $g$ has maximal expressive power, e.g., an MLP, this model can be shown to distinguish all non-isomorphic graphs. Further,~\citet{Ker+2021} studied unique node identifiers in the context of large random graphs.

\paragraph{Randomized Node Labels}
\citet{Mur+2019,Sat+2020,Abb+2021} showed that adding random features, e.g., sampled from the standard uniform distribution, concatenated to the initial node features, enhances the expressive power of GNNs. Specifically,~\citet{Sat+2020} showed that adding random features to the initial features of the GIN layer of~\cref{gin} improves their ability to randomly approximate the solution of common combinatorial optimization problems, e.g., minimum dominating set problem and maximum matching problem, over standard GNNs. \citet{Abb+2021} investigated the universality, see~\cref{approx} below, of such architectures. They showed that adding random features to GNNs results in universality for the class of invariant functions on graphs with high probability. \citet{Das+2020} obtained similar universality results also leveraging random colorings.

\paragraph{Homomorphism- and Subgraph-based Approaches}
\citet{botsas2020improving} extended the expressive power of GNNs by enhancing them with subgraph information. Specifically, they fix a set of small subgraphs $\mathcal{F}$ of given graph $G$. For each node $v$ in $V(G)$ and each subgraph $F$ in $\mathcal{F}$, they compute the node's role in the subgraph and add this information to the node's feature. That is, formally, they compute the automorphism type of node $v$ concerning the subgraph $F$. Similarly, they add information based on the edge automorphism type. Theoretically, they derived conditions under which the enhanced GNNs become more powerful than the $1$-WL, based on the choice of the set of subgraphs $\mathcal{F}$.
By relying on  homomorphism counts,~\citet{Bar+2021} analyzed under which conditions adding more subgraphs leads to added expressivity and studied  the expressive power of the resulting architectures compared to the $k$-FWL. Moreover,~\citet{Hoa+2020} directly leveraged the connection between homomorphism counts and the $k$-FWL hierarchy, see~\cref{theory}, and proved universality results for such architectures.

\paragraph{Subgraph-enhanced Approaches}
Recently, another type of subgraph-based approach to enhance GNNs' expressive power emerged; see, e.g.,~\citet{Bev+2021,Cot+2021,Li+2020,Pap+2021,Thi+2021,You+2020,Wij+2022,Zha+2021b}. These approaches enhanced the expressive power of GNNs by 
representing graphs as multi-sets of subgraphs and applying GNNs to these subgraphs. The subgraphs are obtained by removing, extracting, or marking (small) subgraphs to allow GNNs to leverage more structural patterns within the given graph, essentially breaking symmetries induced by the GNNs' local aggregation function. We henceforth refer to these approaches as \new{subgraph-enhanced GNNs}.

For example, \citet{Cot+2021} derived a more powerful graph representation based on ideas inspired by the graph reconstruction conjecture~\citep{bondymanual}. They showed that removing single vertices and deploying GNNs on the resulting subgraphs leads to more powerful GNN architectures. Moreover, they showed that such architectures can be made more powerful by removing several vertices simultaneously, distinguishing graphs the $2$-FWL cannot. \citet{Pap+2021} proposed a similar approach. Instead of removing all vertices, the authors proposed to remove vertices randomly. \citet{Pap+2022} compared these approaches' expressive power to the subgraph-based approaches; see the previous paragraph.

\citet{You+2020} proposed, for each node $v$, to extract its $k$-disc, i.e., the graph induced by all nodes at a distance at most $k$ from node $v$, and assigned a unique marking to node $v$. Each message passing iteration used two aggregation functions with distinct parameters. One function aggregates features around node $v$ and the other aggregates around all other subgraph nodes. They showed that this architecture can, e.g., count the number of cycles starting at node $v$, predict the clustering coefficient, or distinguish random $d$-regular graphs. Hence, making it strictly more powerful than standard GNNs. \citet{San+2021} enhanced GNN's expressive power by proposing an architecture performing message passing within each node's \new{ego network}, i.e., the subgraph induced by a node and its neighbors, and across ego networks. The authors show that such architecture can distinguish the graphs of~\cref{fig:wlcounter}. Similarly,~\citet{Zha+2021} proposed to make GNNs more powerful by extracting the $k$-hop neighborhood around each node and applying a standard GNN on top. The resulting node representations for each subgraph are then pooled together to learn a single representation for each node. Under certain assumptions, the authors showed that such an architecture can distinguish regular graphs.

Moreover,~\citet{Bev+2021} generalized several ideas discussed above and proposed a framework in which each graph is represented as a subset of its subgraphs and processed using an equivariant architecture based on the Deep Sets for Symmetric elements architecture~\citep{maron2020learning} and message-passing neural networks. The authors showed that several simple subgraph selection policies, e.g., edge removal, ego networks, or node removal, generate more powerful GNNs, and derived equivalent WL-like procedures. In follow-up work,  \citet{frasca2022understanding} presented a novel symmetry analysis for several of the approaches mentioned above \citep{Bev+2021,You+2020,Cot+2021,Zha+2021b}, for the common case in which subgraphs are selected in one-to-one correspondence with nodes (for example, by deletion of nodes, node marking, or extraction of ego networks). Based on this symmetry analysis, they were able to link subgraph-enhanced GNNs with previously studied equivariant models for graphs \citep{Mar+2019c}, thereby defining a systematic framework to develop novel architectures extending this family of architectures, as well as proving an upper bound on the expressive power of these methods by $3$-OWL.

\cite{Qia+2022} introduced a theoretical framework to study and generalize the approaches in the last three paragraphs. They showed that all such subgraph-enhanced approaches with subgraph size bounded by $k$ are limited by the $(k+1)$-FWL while being incomparable to the $k$-FWL in terms of distinguishing non-isomorphic graphs. Moreover, based on~\cite{Nie+2021}, they explored data-driven sampling techniques to select subgraphs. Finally, recently,~\cite{Za+2023} conducted a more fine-grained, general analysis of subgraph-enhanced GNNs. Besides other things, they derived a subgraph-enhanced GNN of maximal expressive power, devised equivalence classes for different types, and developed new theoretical tools for their analysis.  

\paragraph{Other Approaches}

\citet{Toe+2021} proposed an architecture using random walks to extract substructures from a graph. For each node, they uniformly and at random sampled a set of random walks from a graph. They collected features along the walks and constructed a feature matrix processed by 1D convolutions followed by an MLP to update the node's feature. Moreover, they showed under which conditions such architecture exceeds the expressive power of the $k$-FWL. Leveraging the results in~\citet{Cai+1992}, they derived pairs of non-isomorphic graphs the $k$-FWL cannot distinguish, see~\cref{theory}, while their proposed architecture, using walks of length $k^2$ and $\mathcal{O}(n)$ samples, distinguishes them. However, they also derive pairs of graphs that the $1$-WL can distinguish, but their architecture cannot.

\citet{Bod+2021} defined a variant of the Weisfeiler--Leman algorithm for handling \emph{simplicial complexes}---generalizations of graphs, incorporating higher-dimensional connectivity such as cliques. Moreover, they proposed a corresponding neural architecture showing that it is more powerful than the $1$-WL while being able to distinguish graphs the $2$-WL cannot. This extension is seen to substantially improve classification performance at the price of higher memory requirements and increased running time. In~\citet{Bod+2021b}, building on the above, \citet{Bod+2021b} also defined a variant of the Weisfeiler--Leman algorithm for cellular complexes generalizing simplicial complexes.

\citet{Li+2020} enhanced GNNs with distance information, e.g., random walks, and showed under which conditions such additional information leads to more powerful node and graph embeddings than GNNs. Further works overcome $1$-WL limitations by including edge~\citep{Kli+2020}, spectral~\citep{Bal+2021}, and directional information~\citep{beaini2020directional}. A different strategy is adopted by \citet{Horn+2022}, who prove that the integration of low-dimensional topological features~(specifically, connected components and cycles) can be used to develop graph neural networks that are more powerful than the $1$-WL. The use of topological calculations adds an additional complexity factor of $\cO(m \log m)$ to the calculation of $1$-WL features or GNN features, with $m = |E(G)|$.
An extension of this work recently showed that higher-order topological information results in architectures that are at least as powerful as $k$-FWL~\citep{rieck2023expressivity}.

\cite{Za+2023b} studied the $1$-WL and GNNs by showing that they are not able to solve problems related to \new{biconnectivity}~\citep{Bol+2002} and derived a variant of the $1$-WL being able to encode general distance metrics, e.g., the shortest-path distance. Further, they derived a transformer-like architecture~\citep{Mul+2023} to simulate this variant. Additionally, they showed that one of the subgraph-enhanced GNNs by~\cite{Bev+2021} can solve the above problems related to biconnectivity. Finally,~\cite{Kim+2022} devised transformer architectures for graphs~\cite{Mul+2023} that are capable of simulating the $2$-FWL.

\paragraph{Node- and Link Prediction}
The above neural architecture beyond $1$-WL's expressive power mainly dealt with graph-level prediction tasks, e.g., graph classification. However, a few works also use $1$-WL's expressivity as a yardstick to study the expressive power of GNNs for node-level or link prediction. For example,~\cite{Zen+2022} explored extracting a connected subgraph around a node $v$ using hand-crafted heuristics. On top of this subgraph, they used a GNN to compute a vectorial representation or feature for the node $v$. In turn, this feature is used, e.g., to classify the node $v$ in a node classification setting. \citet{Zen+2022} showed that the above method can distinguish nodes in a graph that the $1$-WL cannot distinguish. Further,~\cite{Hu+20222} explored GNNs inspired by the $2$-WL for link prediction.

\section{Equivariant Graph Networks and the Weisfeiler--Leman Algorithm}\label{sec:equiv}

In the following, we give an overview of recent progress in the design of equivariant (higher-order) graph networks and their connection to the Weisfeiler--Leman hierarchy and universality.

\begin{figure}[t]
	\centering
	\includegraphics[height=4cm]{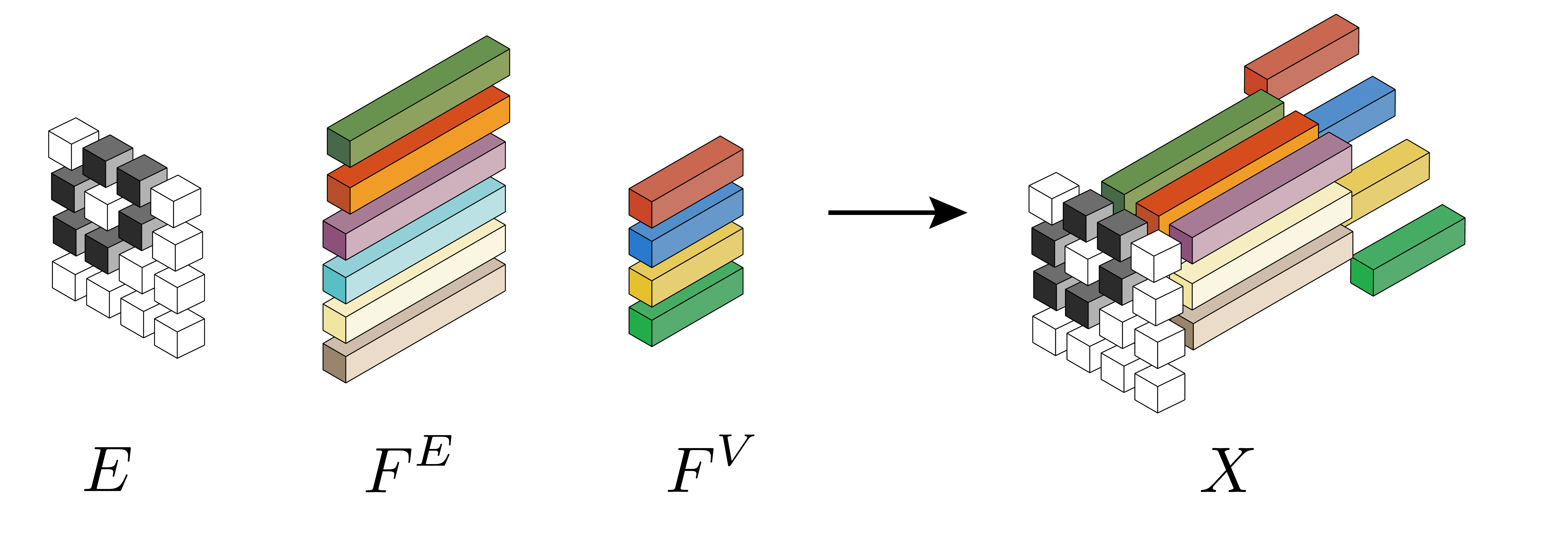}
	\caption{Representing graphs as tensors.  }\label{fig:graphs_as_tensors}
\end{figure}

\subsection{Equivariant Graph Networks}
%\cm{@Haggai, you need adapt use of $k$-WL, $k$-FWL, and $k$-OWL.}

This section shows how graph neural networks can be deduced from the first principles, namely invariance and equivariance to the action of node permutation. We show how these networks, called \emph{Equivariant Graph Networks} (EGN), naturally relate to message-passing GNNs and the Weisfeiler--Leman hierarchy and discuss their expressive power.

\paragraph{Representing Graphs as Tensors} We start by setting up some notation. As before, a graph $G$ is denoted $(V,E)$, with node set $V(G)$ and edge set $E(G)$. We let $n=|V(G)|$ and $m=|E(G)|$ denote the number of nodes and edges, respectively. We further assume that the graph has node features $F^{V}$ in $\mathbb{R}^{n\times d}$ and (potentially) edge features $F^{E}$ in $\mathbb{R}^{m\times d}$. In this section, we encode all the graph data, i.e., adjacency information $E(G)$ and features $F^{V}$, and $F^{E}$ as three-dimensional tensors,
$$X\in \mathbb{R}^{n^2\times (2d+1)}.$$
The first $n^2=n\times n$ slice, namely $X_{:,:,1}$, holds the adjacency matrix of the graph, which is determined by the set of edges $E(G)$. The next $d$ channels $X_{:,:,2:d+1}$ hold the edge features, namely,  $X_{i,j,2:d+1}=F^{E}_{e,:}$, where, with a slight abuse of notation, $e=(i,j)$ in $E$. Similarly the last $d$ channels  $X_{:,:,d+2:2d+1}$ hold the node features on the diagonal, $X_{i,i,d+2:2d+1}=F^{V}_{i,i,:}$, and zeros on the off-diagonals. See Figure \ref{fig:graphs_as_tensors} for an illustration of this construction.

A generalization of the graph tensor representation is
\begin{equation*}
	X \in \mathbb{R}^{n^k \times c},
\end{equation*}
where we attach feature vectors in $\mathbb{R}^c$ to $k$-tuples of nodes. That is, for a $k$-tuple of nodes $\vec{v}=(i_1,i_2,\ldots,i_k)$, for $i_j$ in $[n]$, we attach the feature vector $X_{i_1,i_2,\ldots,i_k,:}$ in  $\mathbb{R}^c$. This representation can be seen as a method of encoding the coloring of the $k$-OWL algorithm  as described in Section \ref{ss:k_dim_WL_algorithm}, where colors are represented as feature vectors. Furthermore, this representation can be used to represent hypergraphs~\citep{Mar+2019c}.

An alternative way of representing graphs as tensors is using incidence matrices as suggested in~\citet{albooyeh2019incidence}. Here a (feature-less) graph is represented as a tensor $X$ in $\mathbb{R}^{n\times m}$, where $X_{i,e}=1$ if the $i$-th node is incident to the edge $e$, and $X_{i,e}=0$ otherwise. Features on nodes or edges can be encoded using extra channels of $X$. Node features $F^V$ can be added as $d$ layers $X_{i,e,2:d+1} = F^V_{i,:}$, for all $e$ in $E$, while edge features $X_{i,e,d+2:2d+1} = F^E_{e,:}$ for all $i$.

\paragraph{Symmetries of Graph Tensor Representations}
Structured objects can often undergo transformations that do not change their essence. Such transformations are called symmetries and are mathematically defined by a group $\mathcal{G}$ that acts on the objects. A well-known example of a symmetry is the translation of an image. Applying a translation to an image does not change the objects appearing in it, assuming they are not taken out of the image boundaries by the translation. Symmetries in graphs arise because the nodes, in most cases, do not follow a canonical order, and any order of the nodes may result in an equivalent yet, seemingly different representation of the graph.  More formally, let $S_n$ denote the group of permutations on $n$ symbols and a graph represented as a tensor $X$ in $\mathbb{R}^{n\times n\times d}$, as defined above. Then for any $\tau$ in $S_n$ a reordered version of the tensor, $\tau\cdot X$, defined by $$(\tau\cdot  X)_{ijk} = X_{\tau^{-1}(i),\tau^{-1}(j),k},$$ represents exactly the same graph. When considering a higher-order tensor representation, the symmetries are defined similarly by $(\tau\cdot X)_{i_1,\ldots,i_k,j} = X_{\tau^{-1}(i_1),\ldots,\tau^{-1}(i_k),j}$.

The incidence tensor representation admits a different symmetry where rows corresponding to nodes and columns corresponding to edges can be  permuted independently. That is, for  $(\tau,\nu)$ in $S_n\times S_n$ the symmetry transformation on an incidence matrix $X$ in $\mathbb{R}^{n\times m}$ can be written by $((\tau,\nu)\cdot X)_{ijk} = X_{\tau^{-1}(i),\nu^{-1}(j),k}$; see \citet{albooyeh2019incidence}. In this manuscript, we will focus on the tensor representation and its generalization due to its tight connection to the Weisfeiler--Leman algorithm.

\paragraph{Equivariance as a Design Principle for Neural Networks}
The vast majority of graph learning tasks belong to one of two groups: invariant or equivariant. In cases where a single output is predicted for the entire graph, e.g., when solving graph classification problems, the output is often \emph{invariant} under the node relabeling operation described above, namely $f(\tau\cdot X)=f(X)$. In other cases, predicting values for every node, e.g., in node classification problems or every edge of a graph, may be required. In these cases, the task is often \emph{equivariant} to the relabeling operation, namely $f(\tau \cdot X)=\tau\cdot  f(X)$.

In learning invariant or equivariant graph functions, restricting the model \emph{by construction} to be invariant or equivariant is often preferable to train more powerful models. For example, recent studies demonstrated that invariant models enjoy better generalization~\citep{Bie+2021,elesedy2021provably,garg2020generalization,liao2020pac,Mei+2021,sokolic2017generalization} and improved efficiency \citep{Mar+2019,Zah+2017}. Indeed, recent years have seen the introduction of equivariance and invariance as a leading design principle for deep learning models for structured data \citep{bronstein2021geometric,cohen2016group,ravanbakhsh2017equivariance,wood1996representation}.

\paragraph{Invariant and Equivariant Architectures}
To practically build equivariant networks, we first need to choose our graph tensor representation, each endowed with its symmetries, as described above. Second, we compose multiple equivariant layers, as follows,
$$f_{\text{equi}}=L_1\circ \dots \circ L_k,$$
where $L_i$ are simple ``primitive'' equivariant functions with tunable parameters. Drawing inspiration from multi-layer perceptrons (MLPs), each $L_i$ can be chosen to be an affine equivariant transformation composed with an entrywise non-linearity, such as the ReLU function.
Since invariant linear transformations are often rather limited,
invariant networks are constructed by composing a single invariant layer, potentially followed by an MLP, to the equivariant architecture,
$$f_{\text{inv}}=L_{\text{inv}}\circ f_{\text{equi}},$$
where $L_{\text{inv}}$ is some simple, potentially tunable, invariant layer. As a consequence of the constructions just described, the problem of constructing equivariant models is reduced to finding simple and powerful primitive invariant, $L_{\text{inv}}$, and equivariant, $L_i$, blocks.

The first paper to consider the construction above for graph learning is the pioneering work of \citet{kondor2018covariant} that suggested a set of linear and non-linear equivariant layers for tensors with $S_n$ symmetry. While not characterizing the full spaces of equivariant layers, the authors identified several important instances: tensor product, tensor contraction, and tensor projection. Since then, a core research theme in this field has become the characterization of useful families of equivariant layers. For the incidence matrix representations, \citet{albooyeh2019incidence} presented the full characterization of affine layers, while the full characterization for the tensor representation was provided in \citet{Mar+2019c}.

\paragraph{Linear Equivariant Layers for Graphs}
We will elaborate on constructing invariant and equivariant linear layers for the graph tensor representation. To keep things focused and concise, we will assume a single feature dimension and no bias; see further information in \cite{Mar+2019}.

A general linear transformation $L:\mathbb{R}^{n^2} \rightarrow \mathbb{R}^{n^2}$ would be equivariant if it satisfies the following system of linear equations
$$ L(\tau \cdot X)=\tau\cdot L(X),\quad \forall X\in \mathbb{R}^{n^2},\tau \in S_n.$$
To characterize the solutions to this system let us represent $L$ as a tensor $L\in \mathbb{R}^{n^4}$, and $L(X)_{i,j} = \sum_{kl}L_{i,j,k,l} X_{k,l}$. Now the above equations take the form: $$\sum_{kl}L_{i,j,\tau(k),\tau(l)} X_{k,l} = \sum_{kl}L_{\tau^{-1}(i),\tau^{-1}(j),k,l} X_{k,l},$$
which holds for all $X$ and $\tau$ in $S_n$ if and only if
$$L_{i,j,k,l}=L_{\tau(i),\tau(j),\tau(k),\tau(l)}, \quad \forall \tau\in S_n.$$
That is, $L$ is equivariant if it is constant on the orbits of the action of $S_n$ on $[n]^4$. Each orbit is characterized by a unique \emph{equality pattern}. For example, all indices $(i_1,i_2,i_3,i_4)\in [n]^4$ such that $i_1=i_2=i_3=i_4$ or all indices such that $i_1=i_2\neq{}i_3=i_4$. A simple counting argument \citep{Mar+2019b} shows that there are exactly $\text{bell}(4)=15$ such equality patterns. Here, $\text{bell}(k)$ is the bell number which counts the number of different partitions of a set of size $k$. Therefore, the space of equivariant linear operators $L \colon \mathbb{R}^{n^2}\to \mathbb{R}^{n^2}$ is spanned by the set of indicator matrices of the equality patterns,
$$E^j\in \mathbb{R}^{n^2\times n^2},\quad j\in [15].$$
For example, if the $j$-th equality pattern is given by  $i_1=i_2=i_3=i_4$ then
$$E^j_{i_1,i_2,i_3,i_4}=
	\begin{cases}
		1 & \text{if } i_1=i_2=i_3=i_4 \\
		0 & \text{otherwise. }         \\
	\end{cases}
$$
In \citet{ravanbakhsh2017equivariance,wood1996representation} this principle is discussed in more general terms. Note, however, that for more general group actions and representations, a closed-form solution and characterization of the space of equivariant linear operators, as done above, might be hard to find or calculate.
To summarize the above discussion, we have the following characterization~\citep{Mar+2019c}.
\begin{theorem}[Characterization of linear graph equivariant layers] \label{thm:char_equivariant_2_2}
	The space of linear $S_n$-equivariant layers  $L \colon \mathbb{R}^{n^2} \to \mathbb{R}^{n^2}$ is $\text{bell}(4) = 15$-dimensional, and can be written as
	$$L=\sum_{j=1}^{15}  w_j E^j,$$ where each $E^j$ in $\mathbb{R}^{n^2\times n^2}$ is an indicator tensor of an equality pattern on $4$ indices. The scalars $w_j$ in $\mathbb{R}$ are the learnable parameters of this layer.
\end{theorem}
Notably, the dimension of the space of equivariant layers is independent of the graph size $n=|V(G)|$. This crucial property allows practitioners to apply these layers to graphs of any size by using the same learned parameters $w_j$ and $E_j$ matrices of appropriate size for each graph. Similar arguments as above lead also to a generalization of Theorem \ref{thm:char_equivariant_2_2} to equivariant linear operators between higher order tensors $L \colon \mathbb{R}^{n^k} \to \mathbb{R}^{n^\ell}$:
\begin{theorem}[Characterization of linear hyper-graph equivariant layers]\label{thm:char_equivariant_n_m}
	The space of linear $S_n$-equivariant layers  $L \colon \mathbb{R}^{n^k} \to \mathbb{R}^{n^\ell}$ is $\text{bell}(k+l)$ dimensional, and can be written as
	$$L=\sum_{j=1}^{\text{bell}(k+l)} w_jE^j,$$ where each $E^j$ in $\mathbb{R}^{n^k\times n^l}$ is an indicator tensor of an equality pattern on $k+l$ indices. As before, the scalars $w_j$ in $\mathbb{R}$ are the learnable parameters of this layer.
\end{theorem}

\paragraph{Non-linear Equivariant Layers for Graphs}
A natural generalization of the linear equivariant layers is non-linear equivariant layers. While such characterizations exist, e.g., for equivariant set polynomials $\mathbb{R}^n\to\mathbb{R}^n$; see for example~\citet{segol2019universal}, until recently, it was not known for the more general case of equivariant graph polynomials $\mathbb{R}^{n^k}\to\mathbb{R}^{n^l}$. 

A set of quadratic equivariant tensor operations are suggested in~\citet{kondor2018covariant} based on tensor arithmetic. One systematic way to achieve quadratic equivariant tensor operators based on the above characterization of linear equivariant operators is by composing a tensor product $X\otimes X$ defined by $$(X\otimes X)_{i_1,\ldots,i_k,j_1,\ldots,j_l} = X_{i_1,\ldots,i_k}X_{j_1,\ldots,j_l},$$ with a general linear equivariant operator $\mathbb{R}^{n^{k+l}}\to \mathbb{R}^{n^m}$ as characterized in~\cref{thm:char_equivariant_n_m}.

Among this myriad of quadratic operators, matrix products have been demonstrated, via relation to Weisfeiler--Leman as described below, to be of particular interest~\citep{Mar+2019}.
Given two matrices $X^1$ and $X^2$ in $\mathbb{R}^{n^2}$,
\begin{equation}\label{e:matrix_product}
	(X^1 X^2)_{i_1,i_2} = \sum_{j=1}^n X^1_{i_1,j}X^2_{j,i_2}.
\end{equation}

Furthermore, beyond quadratic operators, the following generalization of matrix product also found a tight connection to higher-order Weisfeiler--Leman algorithms~\citep{Azi+2020,Mar+2019b}.
Given $X^1,\ldots,X^k$ in $\bbR^{n^k}$, then $\odot_{i=1}^k X^i$ in $\mathbb{R}^{n^k}$ defined by
\begin{equation}\label{e:generalized_matrix_product}
	(\odot_{i=1}^k X^i)_{i_1,\ldots,i_k} = \sum_{j=1}^n X^1_{j,i_2,\ldots,i_k}X^2_{i_1,j,\ldots,i_k}\cdots X^k_{i_1,\ldots,i_{k-1},j} .
\end{equation}

Very recently, \citet{puny2023equivariant} provided a full characterization of equivariant graph polynomials $P \colon \bbR^{n^2} \to \mathbb{R}^{n^2}$. In particular, they presented a basis for the space of equivariant polynomials, where each element, $P_H$, corresponds to a specific multi-graph $H$. These polynomial spaces are used to suggest a new hierarchy for studying the expressive power of GNNs called \emph{polynomial expressiveness}. Lastly, they discuss how to analyze and enhance the polynomial expressiveness of existing GNN models. 

\subsection{Expressive Power and Weisfeiler--Leman Hierarchy}\label{approx}

Here, we analyze the expressive power of the above linear and non-linear equivariant layers in the context of the Weisfeiler--Leman hierarchy.

\paragraph{Separation Power Versus Function Approximation} There are two main notions used to describe the expressive power of  GNNs \citep{chen2019}, graph separation power, and  function approximation power.

Graph separation power refers to the ability of a model class to have an instance that provides a different output on a pair of non-isomorphic graphs $G_1$ and $G_2$. More formally, we say that a parametric model class $\mathcal{C}=\{f(\, \cdot\, ;W) \mid W \in \mathbb{R}^p \}$, where $W$ denotes the model's parameters, can separate a set of graphs $\mathcal{G}$ if for every non-isomorphic $G_1$ and $G_2$ in $\mathcal{G}$ there exist parameters $W$ such that
$$f(G_1;W)\neq f(G_2;W).$$

Function approximation, on the other hand, is the ability of a model class to approximate a given invariant or equivariant function on graphs. Again, more formally, given a compact domain of graphs, $K\subset \mathbb{R}^{n^2\times (2d+1)}$, we say that a parametric model class $C$ can approximate some set of continuous invariant functions over $K$, $\mathcal{F}\subset C(K)$, if for every $g$ in $\mathcal{F}$ and $\epsilon>0$ there exist parameters $W$ in $\mathbb{R}^p$ such that
$$ \|f(\cdot;W)-g(\cdot)\|\leq \epsilon,$$
where $\| \cdot \|$ stands for the $L_{\text{inf}}$ norm on functions: $\|g\|=\max_{x\in K}|g(x)|$.

While seemingly different, \citet{chen2019} proved that these two notions are equivalent. A model class can separate all graphs if and only if it can approximate any continuous invariant function. We shall later see that \citet{Azi+2020} provided a generalization of this result for $k$-FWL and $k$-OWL separation.

To quantify the power of equivariant graph networks, we would like to measure its separation power compared to the Weisfeiler--Leman hierarchy.

\paragraph{Equivariant Networks With Linear Layers} 
We now draw the connection between EGNs with linear equivariant layers to the $k$-OWL algorithm. We will show that EGNs can simulate the $k$-OWL algorithm and, consequently, that they have a separation power of at least the $k$-OWL.

To show that, we need to fix a way to represent colors and multisets of colors. Colors will be represented as vectors in some Euclidean space $\mathbb{R}^d$. We will need the following lemma for representing multisets of colors.
\begin{lemma}\label{lem:multisets}
	Let $\mathbb{R}^d$ be a color space, there exist continuous invariant functions $\phi_j \colon \mathbb{R}^d\to \mathbb{R}$, $j$ in $[J]$, where $J=O((m+d)^d)$, so that a multiset consisting of $n$ elements in $\mathbb{R}^d$, namely $\mathcal{X}=\oms x_i\in\mathbb{R}^{d} \ \vert \ i\in [n] \cms$, is represented uniquely by a vector in $\mathbb{R}^{J}$ defined by $$\Phi(\mathcal{X})_j=\sum_{i=1}^n \phi_j(x_i).$$
\end{lemma}
This lemma was proved, e.g., in \citet{Mar+2019}. The essence is that $\phi_j$ is chosen to be a basis of $d$-multivariate polynomials of degree $n$, and the sum of this basis over the different constituents of $\mathcal{X}$ provides a unique (moment-like) representation to each multiset. Another technical tool we require is a standard approximation power result for MLPs \citep{hornik1991approximation,pinkus1999approximation}.
\begin{theorem}\label{thm:mlp_universal}
	The set of one hidden layer MLPs with a continuous activation $\sigma$, i.e., $\mathcal{M}(\sigma)=\mathrm{span}\{\sigma(w^Tx+b) \ \vert \ w\in\mathbb{R}^n, b\in\mathbb{R} \}$ is dense in $C(\mathbb{R}^n)$ in the topology of uniform convergence over compact sets if and only if $\sigma$ is not a polynomial.
\end{theorem}

Let us consider the multisets $\oms C(\ldots,i_{\ell-1},j,i_{\ell+1},\ldots) \vert \  j\in [n] \cms$ required for implementing the update rule of the  $k$-OWL. Consider the following linear operator $T \colon \mathbb{R}^{n^k}\to \mathbb{R}^{n^k\times k }$
\begin{equation*}
	(TX)_{i_1,\ldots,i_{k},\ell} \mapsto \sum_{j} X_{\ldots,i_{\ell-1},j,i_{\ell+1},\ldots}.
\end{equation*}
This operator is equivariant, as can be verified directly from the definition. In particular, this operator belongs to the linear space characterized in Theorem \ref{thm:char_equivariant_n_m}, i.e., there is a choice of $w_j$ so that $L=T$. Applying $T$ to each feature dimension after applying $x\mapsto (\phi_1(x),\ldots,\phi_J(x))$ to each feature vector of $X$, namely $X_{i_1,\ldots,i_k,:}$ in $\mathbb{R}^d$, encodes the colors. That is, if $Y$ is the output of this operation, then $Y_{i_1,\ldots,i_k,:,:}$ uniquely represents the neighborhood $M^*(i_1,\ldots,i_k)$ of the $k$-OWL algorithm (see Equation \ref{mio}). Using Theorem \ref{thm:mlp_universal} to replace the $\phi_j$ maps with arbitrary good MLP approximations concludes that $f_{\text{equi}}$ equipped with linear equivariant layers as in Theorem \ref{thm:char_equivariant_n_m} can represent the update step of $k$-OWL. Note, however, that tensors of order $k$ are used in the network. We say $f$ is $k$-order if it uses tensors of degrees less or equal to $k$. Consequently, $k$-order $f_{\text{equi}}$ can represent any finite number of $k$-OWL update steps. This is the core argument in the following theorem \citep{Mar+2019b}.
\begin{theorem}\label{thm:separation_k_wl}
	Given two non-isomorphic graphs $G_1$ and $G_2$ that can be distinguished by $k$-OWL algorithm, there exists $k$-order $f_{inv}$ and weights $W$ so that $f_{\text{inv}}(G_1;W)\neq f_{\text{inv}}(G_2;W)$.
\end{theorem}

\paragraph{EGNs with Polynomial Layers and $k$-WL}

The matrix product, \cref{e:matrix_product}, and the generalized matrix product, \cref{e:generalized_matrix_product}, can be used to design equivariant graph networks that can simulate the $k$-FWL algorithm \citep{Azi+2020,Mar+2019b}.
The connection between matrix products and the $k$-FWL test can be illustrated by considering the case $k=2$ studied in \citet{Mar+2019b}, where it was shown that second-order EGNs augmented with a matrix product operation are capable of simulating 2-FWL. In order to demonstrate this, the authors exploit the similarity between the 2-FWL's update rule in Equation \eqref{mi} and the matrix multiplication from~\cref{e:matrix_product}. More specifically, when $k=2$, $M_i$, which represents the neighbor aggregation in the $2$-FWL update rule, takes the form: $M_i(\vec{v}) = \{(C^k_{i}(v,w), C^k_{i}(w,v))   \mid w \in V(G) \}$. Upon closer inspection, it becomes apparent that this is a set of tuples whose indices have a striking similarity to those used in matrix multiplication. %This similarity is then utilized for showing that the 2-FWL update step can be simulated by second-order FEGNs. 
See more details in \cite{Mar+2019c}. This can be generalized to higher-order $k$-EGNs with the generalized matrix product in Equation \eqref{e:generalized_matrix_product} and $k$-FWL.
We name these architectures FEGN, to emphasize the relation to FWL. Similarly, we call EGNs with linear layers OEGN. A summary of the separation results of FEGN and OEGN is provided by~\citet{Azi+2020,Gee+2020b}.
\begin{theorem}
	The separation power of $k$-order OEGN coincides with the separation power of $k$-OWL. The separation power of $k$-order FEGN coincides with that of $k$-FWL.
\end{theorem}

\paragraph{Universality}
Several papers have studied the approximation power of
EGNs rather than their ability to separate non-isomorphic graphs. \citet{Mar+2019c} proved the universality of $G$-invariant networks for any permutation
group $G$ when using (very) high-order tensors. When choosing the appropriate permutation group, this also applies to OEGNs. 
This construction, however, is not feasible as it uses $O(n^4)$ order tensors.
A more efficient, but still unfeasible, construction of universal OEGNs  was presented in \citet{rav+2020}; another proof can be found in \citet{Mae+2019}. \citet{keriven2019,Azi+2020} expanded these results to equivariant functions.

\begin{table}[t]
	\centering
	\begin{tabular}{ccc}
		\toprule
		                                        & $k$-OEGN                       & $k$-FEGN          \\
		\midrule
		Time                                    &
		$\mathcal{O}(n^k\cdot \text{bell}(2k))$ & $\mathcal{O}(n^{k+1} \cdot k)$                     \\
		Space                                   & $\mathcal O(n^k)$              & $\mathcal O(n^k)$ \\
		\bottomrule
	\end{tabular}
 	\caption{Time and space complexity of LEGNs and OEGNs with a constant number of layers.}\label{tab:complexity}
\end{table}

\paragraph{Efficient Implementation}
Thus far, we have mainly discussed the theoretical properties of EGNs. The next step is to discuss how these models can be effectively implemented.
A straightforward way to implement the linear layer used by OEGNs involves constructing matrices with values defined according to equality patterns on the indices, as described above. The main disadvantages of this construction are (1) it scales poorly and is impractical for large values of $n$, and (2) the basis elements lack intuitive meaning. For the $k=2$ case, an alternative was suggested by \citet[Appendix A]{Mar+2019c}. In particular, the authors proposed forming a new basis for the space of linear equivariant functions, in which each basis element is defined as a sum of several previous basis elements. The summands are carefully chosen so that each new basis element represents a simple operation and can be implemented in $O(n^2)$ time  complexity, \emph{without} constructing the aforementioned $n^2\times n^2$ indicator matrix. Here is a partial list of the resulting basis elements: a scaled identity operator for the diagonal or the off-diagonal part of the matrix, a row summation operator that broadcasts the sum to the rows or the columns, and a diagonal summation operator that broadcasts the sum to the diagonal or the off-diagonal. A complete list can be found in \citet{Mar+2019c}.
\cite{albooyeh2019incidence} generalized this idea by showing that for general $k$, all linear equivariant layers can be written as a composition of linear pooling (summation) and broadcasting operators.
A lesser amount of thought was given to implementing the generalized matrix multiplication from Equation \eqref{e:generalized_matrix_product}. To our knowledge, only the case $k=2$ was implemented by standard matrix multiplication \citep{Mar+2019,chen2019,Azi+2020}. Time and space complexity: Table \ref{tab:complexity} summarizes the space and time complexity of OEGNs, employing the layers from Theorem \ref{thm:char_equivariant_n_m}, and FEGNs, using the layers from Equation \eqref{e:generalized_matrix_product}, with a constant number of layers and feature dimensions.

The (local) $k$-OWL-based architectures proposed in~\citet{Morris2020b} are implemented quite differently. Each graph with $n$ nodes simulates the $k$-OWL on an auxiliary graph $G^{\otimes k}$ on $n^k$ vertices defined as follows. The vertex set of $G^{\otimes k}$ is the set $V(G)^k$ of all $k$-tuples in $V(G)^k$. The edge set of $G^{\otimes k}$ is defined as follows. Two $k$-tuples $s$ and $t$ are connected by an edge in $G^{\otimes k}$ 	if they are local $i$-neighbors for $i$ in $[k]$; see~\cref{kwlk}. Such edge is annotated with the label $i$. On top of the auxiliary graph $G^{\otimes k}$, they use a GNN powerful enough to simulate a variant of the $1$-WL taking edge labels into account. \citet{Mor+2019} used a similar strategy for the set-based version of~\cref{kgnns}.

\section{Expressivity and Generalization Abilities of GNNs}

The previous sections show that GNNs' expressive power is reasonably well understood by exploiting their connection to the Weisfeiler--Leman hierarchy. While there exists works upper bounding GNNs' generalization error, e.g.,~\cite{garg2020generalization,liao2020pac,Sca+2018}, these approaches express GNNs' generalization ability using only classical graph parameters, e.g., maximum degree, number of vertices, or edges, which cannot fully capture the complex structure of real-world graphs. Recently,~\cite{Mor+23} made progress connecting GNNs' expressive power and generalization ability via the Weisfeiler--Leman hierarchy. They studied the influence of graph structure and the parameters' encoding lengths on GNNs' generalization by tightly connecting $1$-WL's expressivity and GNNs' Vapnik--Chervonenkis (VC) dimension~\citep{Vap+95}. They derived that GNNs' VC dimension depends tightly on the number of equivalence classes computed by the $1$-WL over a given set of graphs. Moreover, the results easily extend to the $k$-WL and many recent, more expressive GNN extensions. Moreover, they showed that GNNs' VC dimension depends logarithmically on the number of colors computed by the $1$-WL and polynomially on the number of parameters.

\section{Applications}\label{sec:applications}
Structured data is ubiquitous in many disciplines, such as cheminformatics, bioinformatics, neuroscience, natural language processing, social network analysis, and computer vision.
The considered data can typically be represented as graphs, but the modeling is not necessarily unique. For example, in bioinformatics, the nodes of a protein graph may represent the amino acids or the secondary structure elements formed by sequences of amino acids. Likewise, the nodes and edges may be annotated by additional information in the form of attributes.
In general, the used graph model can significantly influence learning methods for graphs.
For the comparison of machine learning methods, several standard benchmark data sets were introduced~\citep{hu2020ogb,Mor+2020}, which contain graphs representing various objects and concepts such as small molecules, proteins, and protein-protein interactions, citation networks, as well as letters, fingerprints, and cuneiform signs.
The considered tasks include node, edge, and graph classification or regression in supervised, semi-supervised, and unsupervised settings.
General-purpose graph learning methods based on or closely related to the Weisfeiler--Leman method are applied in all these domains and settings and have proven to be highly effective~\citep{Kriege2020b,Mor+2020}.

In the following, we review the development in selected application areas and focus on domain-specific constraints, for which Weisfeiler--Leman type algorithms have been adapted. Moreover, we present applications of the Weisfeiler--Leman method in a broader context of machine learning.

\subsection{Computer Programs}
The Weisfeiler--Leman graph kernel was used to determine the similarity of computer programs~\citep{Li+2016}. To this end, either the API calls made by the program or the data flow between procedures is represented by a graph. The nodes are labeled by the procedure's name and the data type, respectively. Using this graph model, the similarity computed by the Weisfeiler--Leman graph kernel can, for example, help to find related source code on internet repositories for reuse.

\citet{Nar+2016} observed that for malware detection, an extended graph model is required. To distinguish regular from malicious code, additional context information is essential, e.g., whether a user is aware or unaware of the execution of the code. This context information was added to the graph model, and the Weisfeiler--Leman graph kernel was adapted to incorporate the additional context labels in the re-labeling procedure.

\subsection{Semantic Web}
Knowledge graphs are a prime example of heterogeneous graphs, in which nodes and edges have different types. For example, nodes may be of type ``person'' or type ``movie,'' each having a different set of attributes. Further, an edge between a person and a movie can, among others, represent an ``is-actor-in'' or ``is-screenwriter-of'' relationship.
De Vries (2013) modified the Weisfeiler--Leman graph kernel for application to general Resource Description Framework (RDF) data.
RDF data consists of subject-predicate-object statements. These form directed multigraphs with node and edge labels when interpreted as edges. Typically, one is interested in the similarity between subgraphs extracted for specific instances. Instead of applying the Weisfeiler--Leman graph kernel to such subgraphs, the Weisfeiler--Leman labels are computed in the whole graph but are distinguished according to their depth regarding instances of interest. By this means, the computation was accelerated compared to the subgraph-based method.

\subsection{Cheminformatics}
As detailed in \cref{sec:rw:mol}, methods that encode the neighborhoods of atoms have been developed in cheminformatics independently of the Weisfeiler--Leman method and were widely used for analyzing molecular data before the Weisfeiler--Leman graph kernel was introduced. Besides parallel developments, research in machine learning with graphs and cheminformatics strongly influenced each other.
Since the development of neural networks for graphs gained momentum in machine learning, new methods were quickly adapted for cheminformatics tasks~\citep{Wieder+2020}.

In the standard graph representation of molecules, the atoms correspond to nodes and chemical bonds to edges. A particularity in cheminformatics is that the properties of interest typically depend on the conformations of molecules, i.e., the geometrical arrangement of the atoms in a 3-dimensional space. Since a molecule can have different conformations, determining the relevant one is often part of the problem. However, this information should be exploited for successful learning when reliable coordinates are available. \citet{Gil+2017} applied GNNs to predict quantum properties and represented 3-dimensional coordinates by adding edges encoding the inter-atom distances to obtain rotation-invariance. \citet{Kli+2020} adapted the message-passing approach of GNNs by incorporating the direction in coordinate space to model angular potentials.

For the prediction of fuel ignition quality, \citet{Schweidtmann+2020} proposed a graph neural network architecture relying on standard molecular graphs annotated with atomic and bond features. The proposed GNN model uses a recurrent neural network architecture with a hierarchical combination of standard and higher-order GNNs to capture the long-range effects of atom groups within a molecule. The method shows competitive performance compared to state-of-the-art domain-specific models.

Apart from the standard graph representation of molecules, further models exist, typically representing groups of connected atoms such as rings by a single node with additional attributes~\citep{Rarey1998,Stiefl2006}.
\citet{Fey2020} proposed a hierarchical architecture where two GNNs run simultaneously on two graph representations at different levels of abstraction, passing messages inside each graph and exchanging messages between the two representations. The combination of a standard molecular graph model with an adaption of a tree-like representation~\citep{Rarey1998} is shown to increase the expressivity of the GNN architecture~\citep{Fey2020}.

\subsection{Computer Vision and Graphics}

Graphs have also been becoming increasingly popular in the computer vision domain for learning on scene graphs~\citep{Rap+2017,Xu+2017}, image keypoints~\citep{Li+2019,Fey+2020,Kriege2018}, superpixels~\citep{Mon+2017}, 3D point clouds~\citep{Qi+2017a} and manifolds~\citep{Fey+2018,Han+2019} where relational learning introduces a critical inductive bias into the model.
Notably, various new GNN variants have been proposed, derived as differentiable versions of the $1$-WL algorithm that can incorporate continuous spatial and semantic information into the neighborhood aggregation phase. However, only a few works make this connection explicit.

For example, \citet{Zah+2017} proposed the popular DeepSets model, e.g., for learning on point clouds, and were the first who studied universality guarantees for fixed-size, countable sets and uncountable sets of invariant deep neural network architectures, pioneering work for the connection of $1$-WL and GNNs.
\citet{Herzig+2018} extends this study and characterizes all permutation invariant architectures applied to predicting scene graphs from images.
Furthermore, \citet{Fey+2020} tackles the task of keypoint matching in natural images by using a  differentiable validator for graph isomorphism based on the $1$-WL heuristic; see also the next section.

\subsection{Graph Matching}
\label{sub:graph_matching_application}

A common principle in comparing graphs is identifying correspondences between their nodes that optimally preserve the edge structure. The problem arises in various domains, and variations have been studied under different terms such as \emph{maximum common subgraph isomorphism}~\citep{Kriege2017b}, \emph{network alignment}~\citep{Zhang2016}, \emph{graph matching}~\citep{Gold1996} or \emph{graph edit distance}~\citep{Sanfeliu1983}, often using different algorithmic approaches. These problems are \cnp-hard, and heuristics are widely applied in practice. A particular simple and efficient approach~\citep{Kri+2019a} applies Weisfeiler--Leman refinement to both graphs to obtain a hierarchy as shown in \cref{fig:assignment}. Then, the nodes of the first graph are assigned to the nodes of the second graph by traversing the hierarchy from the leaves to the root. At each node in the hierarchy, all available nodes are matched, and the remaining nodes are passed to the parent in the hierarchy for matching in a later step. By performing the matching within the color classes using the structural results of~\citet{Arv+2015}, it can be guaranteed that the approach constructs an isomorphism for two isomorphic graphs that are amenable to Weisfeiler--Leman refinement~\citep{Arv+2015,Kri+2019a}. Moreover, the node correspondences obtained by this approach preserve more structural information, i.e., lead to a smaller graph edit distance, on several real-world data sets than comparable heuristics that are computationally more demanding~\citep{Kri+2019a}. \citet{Sto+2019} showed that for graphs representing protein complexes, a similarity measure defined with Weisfeiler--Leman labels highly correlates with the graph edit similarity obtained from a minimal sequence of edit operations.

Recent work by \citet{Bai2019,Bai2020} suggests learning similarities or distances based on graph matching using graph neural networks.
The graph edit distance has been approximated using shared GNNs, where the output is fine-tuned by a (non-differentiable) histogram of correspondence scores~\citep{Bai2019}. In follow-up work, \citet{Bai2020} proposed to order the correspondence matrix in a breadth-first-search fashion and to process it further with the help of traditional CNNs. The approaches are affected by the limited expressivity of the used GNNs and do not provide approximation guarantees. \citet{Qin2020} proposed to learn embeddings for graphs using GNNs reflecting the graph edit distance for similarity search in databases using semantic hashing.

Another recent direction, referred to as \emph{deep graph matching}, uses pairwise node similarities obtained from the node features learned by GNNs~\citep{Fey+2020,Wang2019,Zanfir2018}.
\citet{Fey+2020} proposed a method to iteratively improve the consistency of the similarities in a subsequent second stage via a  differentiable validator for graph isomorphism based on the $1$-WL heuristic. The approach maps node identifiers of one graph according to the current node similarities and distributes them via GNNs synchronously in both graphs. The node similarities are then optimized to reduce the resulting node features' differences. The optimization step resembles classical gradient-based graph matching heuristics, which solve a sequence of linear assignment problems~\citep{Gold1996}.
Among other domains, this method significantly improves the alignment of cross-lingual knowledge bases~\citep{Fey+2020}.

\section{Open Challenges, Limitations, Future Research Directions}\label{sec:challenges}

The present work shows that the Weisfeiler--Leman algorithm, its connection to GNNs, and more powerful equivariant architectures for graphs have led to many meaningful insights, advancing machine learning with graphs and relational structures. However, there remain several open challenges in this area of research, some of which we outline here, along with limitations and  directions for future work.

\subsection{Understanding the Interplay of Expressive Power, Generalization, and Optimization}

Although the results presented here, especially the results of~\cref{approx}, neatly characterize the expressive power of equivariant neural architecture based on the Weisfeiler--Leman method, these universal architectures suffer from an exponential dependence on $k$, e.g., due to operating on $k$-order tensors, making them infeasible for large-scale graphs, resulting in the following challenge.
\begin{oproblem}
	Design provably powerful equivariant neural architectures for graphs that better control the trade-off between expressive power and scalability.
\end{oproblem}

Moreover, the expressivity results are of an existentialist nature. While they show the existence of an architecture's weight assignment, they do not guarantee that standard first-order optimization methods, e.g.,~\citet{Kin+2015}, converge to them. Hence, it is an open challenge to understand the interplay of expressive power and optimization.
\begin{oproblem}
	Understand how first-order optimization methods impact the expressive power of WL-based equivariant architecture for graphs.
\end{oproblem}
Even more, the relationship between \emph{generalization} and \emph{optimization} is understood to a lesser extent. Although some results are shedding some light on the generalization ability using classical tools from learning theory, see, e.g., \citet{garg2020generalization,Kri+2018,liao2020pac,Xu+2021}, there exists little research in understanding how optimization influences generalization and what impact graph structure plays. To the best of our knowledge, only~\cite{Key+2021} tackle this problem, relying on a linearized GNN architecture, and investigate the effect of skip connections, depth, or good label distribution on the convergence during training. Hence, to further investigate this problem, we propose the following open challenge.
\begin{oproblem}
	Understand how first-order optimization methods impact the generalization abilities of WL-based equivariant architectures for graphs.
\end{oproblem}

\subsection{Locality and the Role of Depth}
The number of iterations of the $1$-WL or layers of a GNN architecture is typically selected by cross-validation and often small, e.g., smaller than $5$. For larger values, 1-WL's features become too specific, leading to overfitting for graph kernels. In contrast, under particular assumptions, the GNNs' node features become indistinguishable, a phenomenon referred to as \emph{over-smoothing}~\citep{Liu+20}. Moreover, for GNNs, the \emph{bottleneck problem} refers to the observation that large neighborhoods cannot be accurately represented~\citep{Alon2020}. These problems prevent both methods from capturing global or long-range information. Contrarily, depth, i.e., the number of layers, seems to play a crucial role in the loss landscape of (general) neural networks, e.g.,~\citet{Pog+2019}, resulting in the following challenge.
\begin{oproblem}
	Understand the impact of depth in WL-based equivariant architectures on expressivity, optimization, and generalization, and design architectures that can provably capture long-range dependency.
\end{oproblem}

\subsection{Incorporating Expert Knowledge}

Nowadays, kernels based on the $1$-WL and GNNs are heavily used in life sciences to engineering applications. However, their usage is often ad-hoc, not leveraging crucial expert knowledge. In cheminformatics, for example, information on functional groups or pharmacophore properties is often available, but it is not straightforward to explicitly incorporate it into the $1$-WL and GNN pipeline. Hence, we view the development of mechanisms to include such knowledge as a crucial step in making WL-based learning with graphs more applicable to real-world domains, resulting in the following challenge.
\begin{oproblem}
	Derive a methodology to incorporate expert knowledge, i.e., designing WL-based equivariant architectures that provably capture task-relevant graph structure specified by domain experts.
\end{oproblem}

\subsection{Limitations and Future Research Directions}
While the Weisfeiler--Leman algorithm's connections lead to a better understanding of GNNs, it also has limitations. Since the Weisfeiler--Leman algorithm is purely discrete, it is unclear what it reveals about GNNs' expressivity in the presence of attributed graphs, i.e.,  nodes annotated with a real-valued vector. For example, the presence of additional (real-valued) attributes, which $1$-WL cannot process adequately,  might lead to GNNs distinguishing pairs of nodes that the Weisfeiler--Leman algorithm cannot distinguish. Hence, understanding how to adapt the Weisfeiler--Leman paradigm to this setting remains an open challenge. Moreover, the Weisfeiler--Leman hierarchy might be a too coarse-grained yardstick to understand GNNs' expressivity. For example, in the case of trees already, $1$-WL does not give any insights since it solves the isomorphism problem for trees.

Further, even for more complex graph classes, the algorithm only reveals if a GNN will distinguish non-isomorphic graphs within that class. However, it does not indicate if structurally similar graphs will be mapped to features that are close concerning some distance. Hence, developing a more fine-grained hierarchy might lead to new insights. Finally, the algorithm only captures the expressivity of GNNs following the message-passing framework, see~\cref{eq:gnngeneral}, not of the multitude of spectral GNNs. Hence, understanding how spectral information can enhance or complement the Weisfeiler--Leman algorithm and GNNs is vital, resulting in the following challenge.
\begin{oproblem}
    Understand how the Weisfeiler--Leman algorithm can be used to define a more fine-grained notion of similarity beyond the binary graph isomorphism objective.
\end{oproblem}

\section{Conclusion}
We have provided an overview of the uses of the Weisfeiler--Leman method for machine learning with graphs. To this end, we introduced the $1$-WL and its more powerful generalization, the $k$-dimensional Weisfeiler--Leman algorithm, and outlined its theoretical properties. We then thoroughly surveyed graph kernels based on the Weisfeiler--Leman method. Subsequently, we presented results connecting the $1$-WL and graph neural networks, followed by an overview of neural architecture surpassing the limits of the former. Moreover, we gave an in-depth overview of provably powerful equivariant architectures on graphs and their connection to the $k$-WL and surveyed applications for WL-based machine learning architectures. Finally, we identified open challenges in the field and provided directions for future research.

We hope our survey presents a helpful handbook of graph representation learning methods, perspectives, and limitations and that its insights and principles will help spur novel research results at the intersection of graph theory and machine learning.

\section*{Acknowledgements}

Christopher Morris is partially funded by a DFG Emmy Noether grant (468502433) and  RWTH Junior Principal Investigator Fellowship under Germany's Excellence Strategy. 
Nils M.\ Kriege is supported by the Vienna Science and Technology Fund (WWTF) [10.47379/VRG19009].
Bastian Rieck is supported by the Bavarian state government with
funds from the \emph{Hightech Agenda Bavaria}.

\appendix
\vskip 0.2in\bibliography{bibliography}

\end{document}

%% file: figures/wl.tex
\begin{tikzpicture}

\tikzset{line/.style={draw,thick}}
\tikzset{arrow/.style={line,->,>=stealth}}
\tikzset{node/.style={circle,inner sep=0pt,minimum width=15pt}}
\tikzset{snake/.style={arrow,line width=1.2pt,decorate,decoration={snake,amplitude=1.0,segment length=5,post length=5}}}

\node[line,node,fill=color1] (x1) at (0, 0) {};
\node[line,node,fill=color2] (x2) at (1.5, 0) {};
\node[line,node,fill=color3] (x3) at (3, 0) {};
\node[line,node,fill=color3] (x4) at (1.5, -1.5) {};
\node[line,node,fill=color3] (x5) at (3, -1.5) {};

%\node at (2.25, -0.75) {$c^{(1)}$};

\path[line] (x1) to (x2);
\path[line] (x2) to (x3);
\path[line] (x2) to (x4);
\path[line] (x3) to (x5);
\path[line] (x4) to (x5);

\node[line,node,fill=color4] (y1) at (5.5, 0) {};
\node[line,node,fill=color5] (y2) at (7, 0) {};
\node[line,node,fill=color6] (y3) at (8.5, 0) {};
\node[line,node,fill=color6] (y4) at (7, -1.5) {};
\node[line,node,fill=color7] (y5) at (8.5, -1.5) {};

\path[line] (y1) to (y2);
\path[line] (y2) to (y3);
\path[line] (y2) to (y4);
\path[line] (y3) to (y5);
\path[line] (y4) to (y5);

\path[snake,purple] (x2) to [bend right=10] (y4);
\path[snake,purple] (x4) to [bend right=17] (y4);
\path[snake,purple] (x5) to (y4);

\node[inner sep=0pt,fill=white,opacity=0.8,minimum width=3.4cm,minimum height=0.8cm] at (5, -0.75) {};
\node[inner sep=0pt] at (5, -0.75) {$\textsc{Relabel}\Big( \textcolor{color3}{\blacksquare}, \big\{ \!\! \big\{ \textcolor{color2}{\blacksquare}, \textcolor{color3}{\blacksquare} \big\} \!\! \big\} \hspace{-2pt} \Big)$};

\end{tikzpicture}

%% file: figures/example1.tex
\begin{tikzpicture}

\tikzset{line/.style={draw,thick}}
\tikzset{arrow/.style={line,->,>=stealth}}
\tikzset{node/.style={circle,inner sep=0pt,minimum width=15pt}}

\node[line,node,fill=pink] (x1) at (0, 0.75) {};
\node[line,node,fill=pink] (x2) at (-0.75, -0.75) {};
\node[line,node,fill=pink] (x3) at (0.75, -0.75) {};

\path[line] (x1) to (x2);
\path[line] (x1) to (x3);
\path[line] (x2) to (x3);

\node[line,node,fill=pink] (x1) at (2.25, 0.75) {};
\node[line,node,fill=pink] (x2) at (1.5, -0.75) {};
\node[line,node,fill=pink] (x3) at (3.0, -0.75) {};

\path[line] (x1) to (x2);
\path[line] (x1) to (x3);
\path[line] (x2) to (x3);

\node[line,node,fill=green] (x1) at (3.75, 0) {};
\node[line,node,fill=green] (x2) at (4.25, 0.75) {};
\node[line,node,fill=green] (x3) at (5.25, 0.75) {};
\node[line,node,fill=green] (x4) at (5.75, 0) {};
\node[line,node,fill=green] (x5) at (5.25, -0.75) {};
\node[line,node,fill=green] (x6) at (4.25, -0.75) {};

\path[line] (x1) to (x2);
\path[line] (x2) to (x3);
\path[line] (x3) to (x4);
\path[line] (x4) to (x5);
\path[line] (x5) to (x6);
\path[line] (x6) to (x1);

\end{tikzpicture}

%% file: figures/example2.tex
\begin{tikzpicture}

\tikzset{line/.style={draw,thick}}
\tikzset{arrow/.style={line,->,>=stealth}}
\tikzset{node/.style={circle,inner sep=0pt,minimum width=15pt}}

\node[line,node,fill=pink] (x1)  at (0.75,  0.00) {};
\node[line,node,fill=pink] (x2)  at (1.25,  0.75) {};
\node[line,node,fill=pink] (x3)  at (2.25,  0.75) {};
\node[line,node,fill=pink] (x4)  at (2.75,  0.00) {};
\node[line,node,fill=pink] (x5)  at (2.25, -0.75) {};
\node[line,node,fill=pink] (x6)  at (1.25, -0.75) {};
\node[line,node,fill=pink] (x7)  at (3.75,  0.00) {};
\node[line,node,fill=pink] (x8)  at (4.25,  0.75) {};
\node[line,node,fill=pink] (x9)  at (5.25,  0.75) {};
\node[line,node,fill=pink] (x10) at (5.75,  0.00) {};
\node[line,node,fill=pink] (x11) at (5.25, -0.75) {};
\node[line,node,fill=pink] (x12) at (4.25, -0.75) {};

\path[line] (x1) to (x2);
\path[line] (x2) to (x3);
\path[line] (x3) to (x4);
\path[line] (x4) to (x5);
\path[line] (x5) to (x6);
\path[line] (x6) to (x1);
\path[line] (x4) to (x7);
\path[line] (x7) to (x8);
\path[line] (x8) to (x9);
\path[line] (x9) to (x10);
\path[line] (x10) to (x11);
\path[line] (x11) to (x12);
\path[line] (x12) to (x7);

\node[line,node,fill=green] (x1)  at (6.75,   0.00) {};
\node[line,node,fill=green] (x2)  at (7.25,   0.75) {};
\node[line,node,fill=green] (x3)  at (8.25,   0.75) {};
\node[line,node,fill=green] (x4)  at (9.25,   0.50) {};
\node[line,node,fill=green] (x5)  at (9.25,  -0.50) {};
\node[line,node,fill=green] (x6)  at (8.25,  -0.75) {};
\node[line,node,fill=green] (x7)  at (7.25,  -0.75) {};
\node[line,node,fill=green] (x8)  at (10.25,  0.75) {};
\node[line,node,fill=green] (x9)  at (11.25,  0.75) {};
\node[line,node,fill=green] (x10) at (11.75,  0.00) {};
\node[line,node,fill=green] (x11) at (11.25, -0.75) {};
\node[line,node,fill=green] (x12) at (10.25, -0.75) {};

\path[line] (x1) to (x2);
\path[line] (x2) to (x3);
\path[line] (x3) to (x4);
\path[line] (x4) to (x5);
\path[line] (x5) to (x6);
\path[line] (x6) to (x7);
\path[line] (x7) to (x1);
\path[line] (x4) to (x8);
\path[line] (x8) to (x9);
\path[line] (x9) to (x10);
\path[line] (x10) to (x11);
\path[line] (x11) to (x12);
\path[line] (x12) to (x5);

\end{tikzpicture}

%% file: figures/kwl.tex
\begin{tikzpicture}

\definecolor{purple}{RGB}{147,7,204}
\definecolor{green}{RGB}{124,216,143}
\definecolor{blue}{RGB}{10,153,201}
\definecolor{red}{RGB}{213,42,45}
\definecolor{orange}{RGB}{254,128,41}
\definecolor{gray}{RGB}{120,120,120}

\def\border{22pt}

\tikzset{line/.style={draw,line width=1pt}}
\tikzset{node/.style={circle,inner sep=0pt,minimum width=15pt}}

\tikzset{clusternode/.style={shape=circle,inner sep=0pt,minimum width=\border}}
\tikzset{clusterline/.style={draw,line width=\border}}

\begin{scope}[transparency group,opacity=0.4]
  \node[clusternode,fill=blue] (c1) at (0.0, 0.0) {};
  \node[clusternode,fill=blue] (c2) at (2.0, 0.0) {};
  \node[clusternode,fill=blue] (c3) at (1.0, 1.5) {};
  \path[clusterline,blue] (c1.center) to (c2.center);
  \path[clusterline,blue] (c1.center) to (c3.center);
  \path[clusterline,blue] (c2.center) to (c3.center);
  \fill[blue] (c1.center) -- (c2.center) -- (c3.center) -- cycle;
\end{scope}

\begin{scope}[transparency group,opacity=0.4]
  \node[clusternode,fill=orange] (c1) at (1.0, 1.5) {};
  \node[clusternode,fill=orange] (c2) at (3.0, 1.5) {};
  \node[clusternode,fill=orange] (c3) at (2.0, 0.0) {};
  \path[clusterline,orange] (c1.center) to (c2.center);
  \path[clusterline,orange] (c1.center) to (c3.center);
  \path[clusterline,orange] (c2.center) to (c3.center);
  \fill[orange] (c1.center) -- (c2.center) -- (c3.center) -- cycle;
\end{scope}

\begin{scope}[transparency group,opacity=0.4]
  \node[clusternode,fill=green] (c1) at (2.0, 0.0) {};
  \node[clusternode,fill=green] (c2) at (4.0, 0.0) {};
  \node[clusternode,fill=green] (c3) at (3.0, 1.5) {};
  \path[clusterline,green] (c1.center) to (c2.center);
  \path[clusterline,green] (c1.center) to (c3.center);
  \path[clusterline,green] (c2.center) to (c3.center);
  \fill[green] (c1.center) -- (c2.center) -- (c3.center) -- cycle;
\end{scope}

\begin{scope}[transparency group,opacity=0.4]
  \node[node,fill=white] at (0.0, 0.0) {};
  \node[node,fill=white] at (2.0, 0.0) {};
  \node[node,fill=white] at (4.0, 0.0) {};
  \node[node,fill=white] at (1.0, 1.5) {};
  \node[node,fill=white] at (3.0, 1.5) {};
\end{scope}

\node[line,node,label={[yshift=-1.15cm]:$v$}] (x1) at (0.0, 0.0) {$2$};
\node[line,node,label={[yshift=-1.15cm]:$c$}] (x2) at (2.0, 0.0) {$3$};
\node[line,node,label={[yshift=-1.15cm]:$w$}] (x3) at (4.0, 0.0) {$1$};
\node[line,node,label={[yshift=0.1cm]:$a$}] (x4) at (1.0, 1.5) {$1$};
\node[line,node,label={[yshift=0.1cm]:$b$}] (x5) at (3.0, 1.5) {$2$};

\path[line] (x1) to (x4);
\path[line] (x1) to (x5);
\path[line] (x2) to (x3);
\path[line] (x2) to (x4);
\path[line] (x2) to (x5);

\end{tikzpicture}

%% file: figures/example3.tex
\begin{tikzpicture}

\tikzset{line/.style={draw,thick}}
\tikzset{arrow/.style={line,->,>=stealth}}
\tikzset{node/.style={circle,inner sep=0pt,minimum width=11pt}}

  \foreach \i in {0,...,3}{
   \foreach \j in {0,...,3}{
    \node[line,node,fill=pink] (v\i\j) at ($(90*\i:0.8) + (90*\i+35*\j-52.5:1.2)$) {};
   }
  }
  
  \foreach \i in {0,...,3}{
   \foreach \j/\k in {0/1,0/2,0/3,1/2,1/3,2/3}
   \path[line] (v\i\j) edge (v\i\k);
  }
  
  \foreach \i in {0,...,3}{
   \foreach \j/\k in {0/1,0/2,0/3,1/2,1/3,2/3}
   \path[line] (v\j\i) edge (v\k\i);
  }
  
  \foreach \i in {0,...,7}{
   \node[line,node,fill=green] (u\i) at ($(6,0)+(22.5+45*\i:1)$) {};
   \node[line,node,fill=green] (v\i) at ($(6,0)+(22.5+45*\i:2)$) {};
  }
  \foreach \i/\j in {0/2,0/3,0/5,0/6,1/3,1/4,1/6,1/7,2/4,2/5,2/7,3/5,3/6,4/6,4/7,5/7}{
   \path[line] (u\i) edge (u\j);
  }
  \foreach \i/\j in {0/1,0/7,1/0,1/2,2/1,2/3,3/2,3/4,4/3,4/5,5/4,5/6,6/5,6/7,7/0,7/6}{
   \path[line] (v\i) edge (u\j);
  }
  \foreach \i/\j in {0/1,0/2,0/6,0/7,1/2,1/3,1/7,2/3,2/4,3/4,3/5,4/5,4/6,5/6,5/7,6/7}{
   \path[line] (v\i) edge (v\j);
  }
 \end{tikzpicture}

%% file: figures/histogram1.tex
\begin{tikzpicture}[scale=0.8,auto]

\tikzset{edge/.style={draw,thick}}
\tikzset{vertex/.style={edge,circle,inner sep=0pt,minimum width=15pt}}

\colorlet{it0}{gray}
\colorlet{it11}{green}
\colorlet{it12}{red}
\colorlet{it21}{blue!50!white}
\colorlet{it22}{violet}
\colorlet{it23}{orange}
\colorlet{it31}{green!50!black}
\colorlet{it32}{yellow}
\colorlet{it33}{brown}
\colorlet{it34}{blue!50!black}

\node[vertex,fill=it31] (a) at (0,0) {};
\node[vertex,fill=it21,scale=.6] (a1) at (0-.4,0) {};
\node[vertex,fill=it11,scale=.6] (a2) at (0-.7,0) {};
\node[vertex,fill=it0,scale=.6] (a3) at (0-1,0) {};

\node[vertex,fill=it31] (b) at (0,2) {};
\node[vertex,fill=it21,scale=.6] (b1) at (0-.4,2) {};
\node[vertex,fill=it11,scale=.6] (b2) at (0-.7,2) {};
\node[vertex,fill=it0,scale=.6] (b3) at (0-1,2) {};

\node[vertex,fill=it32] (e) at (1.4,1) {};
\node[vertex,fill=it22,scale=.6] (e1) at (1.4-.4,1) {};
\node[vertex,fill=it12,scale=.6] (e2) at (1.4-.7,1) {};
\node[vertex,fill=it0,scale=.6] (e3) at (1.4-1,1) {};

\node[vertex,fill=it33] (c) at (2.8,0) {};
\node[vertex,fill=it21,scale=.6] (c1) at (2.8-.4,0) {};
\node[vertex,fill=it11,scale=.6] (c2) at (2.8-.7,0) {};
\node[vertex,fill=it0,scale=.6] (c3) at (2.8-1,0) {};

\node[vertex,fill=it33] (d) at (2.8,2) {};
\node[vertex,fill=it21,scale=.6] (d1) at (2.8-.4,2) {};
\node[vertex,fill=it11,scale=.6] (d2) at (2.8-.7,2) {};
\node[vertex,fill=it0,scale=.6] (d3) at (2.8-1,2) {};

\node[vertex,fill=it34] (f) at (4.2,1) {};
\node[vertex,fill=it23,scale=.6] (f1) at (4.2-.4,1) {};
\node[vertex,fill=it11,scale=.6] (f2) at (4.2-.7,1) {};
\node[vertex,fill=it0,scale=.6] (f3) at (4.2-1,1) {};

\draw[edge] (a) -- (b); 
\draw[edge] (a) -- (e); 
\draw[edge] (b) -- (e); 
\draw[edge] (e) -- (d); 
\draw[edge] (e) -- (c); 
\draw[edge] (c) -- (f); 
\draw[edge] (d) -- (f); 

\end{tikzpicture}

%% file: figures/histogram2.tex
\begin{tikzpicture}[scale=0.8,auto]

\def\width{10pt}
\def\height{10pt}
\def\offset{15pt}

\tikzset{line/.style={draw,thick}}
\tikzset{arrow/.style={line,->,>=stealth}}
\tikzset{rect/.style={rectangle,inner sep=0pt,minimum width=\width,anchor=south}}

\colorlet{it0}{gray}
\colorlet{it11}{green}
\colorlet{it12}{red}
\colorlet{it21}{blue!50!white}
\colorlet{it22}{violet}
\colorlet{it23}{orange}
\colorlet{it31}{green!50!black}
\colorlet{it32}{yellow}
\colorlet{it33}{brown}
\colorlet{it34}{blue!50!black}

\node[rect,minimum height=6*\height,line,black!20!white,fill=it0] at (0*\offset,0) {};
\node[rect,minimum height=5*\height,fill=it11] at (1*\offset,0) {};
\node[rect,minimum height=1*\height,fill=it12] at (2*\offset,0) {};
\node[rect,minimum height=4*\height,fill=it21] at (3*\offset,0) {};
\node[rect,minimum height=1*\height,fill=it22] at (4*\offset,0) {};
\node[rect,minimum height=1*\height,fill=it23] at (5*\offset,0) {};
\node[rect,minimum height=2*\height,fill=it31] at (6*\offset,0) {};
\node[rect,minimum height=1*\height,fill=it32] at (7*\offset,0) {};
\node[rect,minimum height=2*\height,fill=it33] at (8*\offset,0) {};
\node[rect,minimum height=1*\height,fill=it34] at (9*\offset,0) {};

\draw[line,arrow] (-0.75*\offset,0) -- (10*\offset,0);
\draw[line,arrow] (-0.75*\offset,0) -- (-0.75*\offset,8*\height);

\end{tikzpicture}

%% file: figures/assignment_wl.tex
\begin{tikzpicture}[scale=0.8,auto]

\tikzset{edge/.style={draw,thick}}
\tikzset{vertex/.style={edge,circle,inner sep=0pt,minimum width=15pt}}

\colorlet{it0}{gray}
\colorlet{it11}{green}
\colorlet{it12}{red}
\colorlet{it21}{blue!50!white}
\colorlet{it22}{violet}
\colorlet{it23}{orange}
\colorlet{it31}{green!50!black}
\colorlet{it32}{yellow}
\colorlet{it33}{brown}
\colorlet{it34}{blue!50!black}

\node[vertex,fill=it31, label ={below:$a$}] (a) at (0,0) {};
\node[vertex,fill=it21,scale=.6] (a1) at (0-.4,0) {};
\node[vertex,fill=it11,scale=.6] (a2) at (0-.7,0) {};
\node[vertex,fill=it0,scale=.6] (a3) at (0-1,0) {};

\node[vertex,fill=it31, label ={above:$b$}] (b) at (0,2) {};
\node[vertex,fill=it21,scale=.6] (b1) at (0-.4,2) {};
\node[vertex,fill=it11,scale=.6] (b2) at (0-.7,2) {};
\node[vertex,fill=it0,scale=.6] (b3) at (0-1,2) {};

\node[vertex,fill=it32, label ={above:$e$}] (e) at (1.4,1) {};
\node[vertex,fill=it22,scale=.6] (e1) at (1.4-.4,1) {};
\node[vertex,fill=it12,scale=.6] (e2) at (1.4-.7,1) {};
\node[vertex,fill=it0,scale=.6] (e3) at (1.4-1,1) {};

\node[vertex,fill=it33, label ={below:$c$}] (c) at (2.8,0) {};
\node[vertex,fill=it21,scale=.6] (c1) at (2.8-.4,0) {};
\node[vertex,fill=it11,scale=.6] (c2) at (2.8-.7,0) {};
\node[vertex,fill=it0,scale=.6] (c3) at (2.8-1,0) {};

\node[vertex,fill=it33, label ={above:$d$}] (d) at (2.8,2) {};
\node[vertex,fill=it21,scale=.6] (d1) at (2.8-.4,2) {};
\node[vertex,fill=it11,scale=.6] (d2) at (2.8-.7,2) {};
\node[vertex,fill=it0,scale=.6] (d3) at (2.8-1,2) {};

\node[vertex,fill=it34, label ={right:$f$}] (f) at (4.2,1) {};
\node[vertex,fill=it23,scale=.6] (f1) at (4.2-.4,1) {};
\node[vertex,fill=it11,scale=.6] (f2) at (4.2-.7,1) {};
\node[vertex,fill=it0,scale=.6] (f3) at (4.2-1,1) {};

\draw[edge] (a) -- (b); 
\draw[edge] (a) -- (e); 
\draw[edge] (b) -- (e); 
\draw[edge] (e) -- (d); 
\draw[edge] (e) -- (c); 
\draw[edge] (c) -- (f); 
\draw[edge] (d) -- (f); 

\end{tikzpicture}

%% file: figures/assignment_wl_hierarchy.tex
\begin{tikzpicture}[scale=0.45,auto]

\tikzset{edge/.style={draw,thick}}
\tikzset{vertex/.style={edge,circle,inner sep=0pt,minimum width=13pt}}

\colorlet{it0}{gray}
\colorlet{it11}{green}
\colorlet{it12}{red}
\colorlet{it21}{blue!50!white}
\colorlet{it22}{violet}
\colorlet{it23}{orange}
\colorlet{it31}{green!50!black}
\colorlet{it32}{yellow}
\colorlet{it33}{brown}
\colorlet{it34}{blue!50!black}

\draw[edge,black!50!white,dashed] (-1,0) node[left]{$3$} -- (6.9, 0);
\draw[edge,black!50!white,dashed] (-1,1.4) node[left]{$2$} -- (6.9, 1.4);
\draw[edge,black!50!white,dashed] (-1,2.8) node[left]{$1$}-- (6.9, 2.8);
\draw[edge,black!50!white,dashed] (-1,4.2) node[left]{$i=0$}-- (6.9, 4.2);

\node[vertex,fill=it31, label={below:$\{a,b\}$}] (ab) at (0,0) {};
\node[vertex,fill=it33, label ={below:$\{c,d\}$}] (cd) at (2,0) {};
\node[vertex,fill=it34, label ={below:$\{f\}$}] (f) at (4,0) {};
\node[vertex,fill=it32, label={below:$\{e\}$}] (e) at (6,0) {};
\node[vertex,fill=it21] (abcd) at (1,1.4) {};
\node[vertex,fill=it23] (f2) at (4,1.4) {};
\node[vertex,fill=it22] (e2) at (6,1.4) {};
\node[vertex,fill=it11] (abcdf) at (2.5,2.8) {};
\node[vertex,fill=it12] (e3) at (6,2.8) {};
\node[vertex,fill=it0] (abcdef) at (4.5,4.2) {};

\draw[edge] (ab) -- node {} (abcd); 
\draw[edge] (cd) -- node {} (abcd);
\draw[edge] (f) -- node {} (f2);
\draw[edge] (e) -- node {} (e2);
\draw[edge] (abcd) -- node {} (abcdf);
\draw[edge] (f2) -- node {} (abcdf);
\draw[edge] (e2) -- node {} (e3);
\draw[edge] (abcdf) -- node {} (abcdef);
\draw[edge] (e3) -- node {} (abcdef);

\end{tikzpicture}

%% file: figures/p-wl.tex
\begin{tikzpicture}[start chain,>=latex,node distance=0pt]

\tikzset{line/.style={draw,thick}}
\tikzset{arrow/.style={line,->,>=stealth}}
\tikzset{node/.style={circle,inner sep=0pt,minimum width=15pt}}

	\def\rectWidth{2.1cm}
	\def\rectHeight{1.9cm}
	\def\panelShift{2.0cm}
	\def\labelYShift{-2pt}
	\def\CNodeXShift{5}
	\def\ArrowXShift{10}
	\def\NodeLabelFontSize{\footnotesize}
	{ [start chain=trunk going right]
		% First panel
		\node[fill=white, rectangle, on chain, minimum width=\rectWidth, minimum height=\rectHeight] (firstRect) {};
        \node[line,node,fill=color2] (A) at  (firstRect.north east) {\NodeLabelFontSize A};
		\node[line,node,fill=color1] (B) at  (firstRect.north) {\NodeLabelFontSize B};
		\node[line,node,fill=color3] (C) at  ([xshift=\CNodeXShift]firstRect.west) {\NodeLabelFontSize C};
		\node[line,node,fill=color2] (Ar) at ([xshift=-\CNodeXShift]firstRect.east) {\NodeLabelFontSize A};
		\node (label1) at ([yshift=\labelYShift]firstRect.south) {Input graph $G$};
		% Second panel
		\node[fill=white,rectangle,on chain, minimum width=\rectWidth, minimum height=\rectHeight, xshift=\panelShift] (secondRect) {};
        \node[line,node,fill=color2,label={north:\small$\oms B \cms$}] (A2) at (secondRect.north east) {\NodeLabelFontSize A};
        \node[line,node,fill=color1,label={north:\small$\oms A, A, C \cms$}] (B2) at (secondRect.north) {\NodeLabelFontSize B};
        \node[line,node,fill=color3,label={south:\small$\oms B \cms$}] (C2) at ([xshift=\CNodeXShift]secondRect.west) {\NodeLabelFontSize C};	
        \node[line,node,fill=color2,label={south:\small$\oms B \cms$}] (Ar2) at ([xshift=-\CNodeXShift]secondRect.east) {\NodeLabelFontSize A};
		\node (label2) at ([yshift=\labelYShift]secondRect.south) {Multiset labels};
		% % Third panel
		\node[fill=white,rectangle,on chain, minimum width=\rectWidth, minimum height=\rectHeight, xshift=\panelShift] (thirdRect) {};
         \node[line,node,fill=color2] (A3)  at (thirdRect.north east) {\NodeLabelFontSize X};
		\node[line,node,fill=color1] (B3)  at (thirdRect.north) {\NodeLabelFontSize Y};
		\node[line,node,fill=color3] (C3)  at ([xshift=\CNodeXShift]thirdRect.west) {\NodeLabelFontSize Z};
		\node[line,node,fill=color2] (Ar3) at ([xshift=-\CNodeXShift]thirdRect.east) {\NodeLabelFontSize X};
		\node (label3) at ([yshift=\labelYShift]thirdRect.south) {Weighted graph $G'$};
		% % Fourth panel
		\node[fill=white, rectangle,on chain, minimum width=\rectWidth, minimum height=\rectHeight, xshift=\panelShift] (fourthRect) {};
		% % Bar code
		\draw[fill=color2, draw=none] ([yshift=-0.225cm]fourthRect.west) rectangle ([xshift=0.8cm,yshift=-0.025cm]fourthRect);
		\draw[fill=color1, draw=none] ([yshift=0.075cm]fourthRect.west) rectangle ([xshift=0.3cm,yshift=0.275cm]fourthRect);
		\draw[fill=color3, draw=none] ([yshift=0.375cm]fourthRect.west)  rectangle ([xshift=0.5cm,yshift=0.575cm]fourthRect);
		\draw[fill=color2, draw=none] ([yshift=0.675cm]fourthRect.west)  rectangle ([xshift=-0.2cm,yshift=0.875cm]fourthRect);
		\draw[line,arrow] ([yshift=-0.4cm]fourthRect.west) -- ([yshift=1.15cm]fourthRect.west);
		\draw[line,arrow] ([yshift=-0.4cm]fourthRect.west) -- ([yshift=-0.4cm]fourthRect.east);
		\node (label4) at ([yshift=\labelYShift]fourthRect.south) {Topological relevance};
	}
	% % the arrows and labels
	\draw[line,arrow] ([xshift=\ArrowXShift]firstRect.east) -- ([xshift=-\ArrowXShift]secondRect.west) node[midway,above] {$1$-WL};
	\draw[line,arrow] ([xshift=\ArrowXShift]secondRect.east) -- ([xshift=-\ArrowXShift]thirdRect.west) node[midway,above] {Metric};
	\draw[line,arrow] ([xshift=\ArrowXShift]thirdRect.east) -- ([xshift=-\ArrowXShift]fourthRect.west) node[midway,above] {TDA};
	
	\draw[line] (B.east) -- (A.west) node[midway,above] {};
	\draw[line] (B.south west) -- (C.north east);
	\draw[line] (B.south east) -- (Ar.north west) node[midway,above] {};
	
	\draw[line] (B2.east) -- (A2.west) node[midway,above] {};
	\draw[line] (B2.south west) -- (C2.north east);
	\draw[line] (B2.south east) -- (Ar2.north west) node[midway,above] {};
	
	\draw[line] (B3.east) -- (A3.west) node[midway,above] {$w_1$};
	\draw[line] (B3.south west) -- (C3.north east) node[midway,left] {$w_2$};
	\draw[line] (B3.south east) -- (Ar3.north west) node[midway,right] {$w_3$};
\end{tikzpicture}

%% file: figures/gnn.tex
\begin{tikzpicture}

\tikzset{line/.style={draw,thick}}
\tikzset{arrow/.style={line,->,>=stealth}}
\tikzset{node/.style={circle,inner sep=0pt,minimum width=15pt}}
\tikzset{snake/.style={arrow,line width=1.2pt,decorate,decoration={snake,amplitude=1.0,segment length=5,post length=5}}}

% Bar: id/height1/height2
\tikzset{pics/bar/.style args={#1/#2/#3}{code={%
  \node[inner sep=0pt,minimum height=0.3cm] (#1) at (0,-0.15) {};
  \draw[fill=blue,blue] (-0.15,-0.2) rectangle (-0.075,0.25*#2-0.2);
  \draw[fill=orange,orange] (-0.025,-0.2) rectangle (0.05,0.25*#3-0.2);
  \draw[arrow,semithick] (-0.2,-0.2) -- (0.2,-0.2);
  \draw[arrow,semithick] (-0.2,-0.21) -- (-0.2,0.2);
}}}

\node[line,node] (x1) at (0, 0) {};
\node[line,node] (x2) at (1.5, 0) {};
\node[line,node] (x3) at (3, 0) {};
\node[line,node] (x4) at (1.5, -1.5) {};
\node[line,node] (x5) at (3, -1.5) {};

\pic[] at (0,0.6) {bar=x1-bar/1.0/0.2};
\pic[] at (1.5,0.6) {bar=x2-bar/0.4/1.0};
\pic[] at (3,0.6) {bar=x3-bar/0.7/0.8};
\pic[] at (1.5,-2.1) {bar=x4-bar/0.7/0.8};
\pic[] at (3,-2.1) {bar=x5-bar/0.7/0.8};

\path[line] (x1) to (x2);
\path[line] (x2) to (x3);
\path[line] (x2) to (x4);
\path[line] (x3) to (x5);
\path[line] (x4) to (x5);

\node[line,node] (y1) at (7.5, 0) {};
\node[line,node] (y2) at (9, 0) {};
\node[line,node] (y3) at (10.5, 0) {};
\node[line,node] (y4) at (9, -1.5) {};
\node[line,node] (y5) at (10.5, -1.5) {};

\path[line] (y1) to (y2);
\path[line] (y2) to (y3);
\path[line] (y2) to (y4);
\path[line] (y3) to (y5);
\path[line] (y4) to (y5);

\pic[] at (7.5,0.6) {bar=y1-bar/1.0/0.3};
\pic[] at (9,0.6) {bar=y2-bar/0.2/0.9};
\pic[] at (10.5,0.6) {bar=y3-bar/0.5/0.4};
\pic[] at (9,-2.1) {bar=y4-bar/0.5/0.4};
\pic[] at (10.5,-2.1) {bar=y5-bar/0.7/0.8};

\path[snake,purple] (x2) to [bend right=7] (y4);
\path[snake,purple] (x4) to [bend right=15] (y4);
\path[snake,purple] (x5) to (y4);

\node[inner sep=0pt,fill=white,opacity=0.8,minimum width=5.2cm,minimum height=0.8cm] at (6, -0.75) {};
\node at (6, -0.75) {$ \color{purple!50} f^{W_1}_{\text{merge}}\Big( \hspace{20pt} ,f^{W_2}_{\text{aggr}}\big(\{ \!\! \{ \hspace{20pt}, \hspace{20pt} \} \!\! \} \big)\!\Big)$};

\pic[] at (4.9,-0.75) {bar=f4-bar/0.7/0.8};
\pic[] at (6.8,-0.75) {bar=x2-bar/0.4/1.0};
\pic[] at (7.7,-0.75) {bar=f5-bar/0.7/0.8};

\end{tikzpicture}